\pdfoutput=1

\PassOptionsToPackage{dvipsnames,table}{xcolor}

\documentclass[11pt]{article}

\usepackage[preprint]{acl}

\usepackage[dvipsnames,table]{xcolor}
\usepackage{times}
\usepackage{latexsym}
\usepackage{amsmath}
\usepackage{threeparttable}
\usepackage{tabularx}
\usepackage{array}
\usepackage[T1]{fontenc}
\usepackage{booktabs}
\usepackage{multirow}
\usepackage{tcolorbox}
\tcbuselibrary{skins}
\usepackage{enumitem}
\usepackage{url}
\usepackage{pifont}
\newcommand{\cmark}{\ding{51}}
\newcommand{\xmark}{\ding{55}}
\usepackage[linesnumbered,ruled,vlined]{algorithm2e}
\usepackage[utf8]{inputenc}
\usepackage{microtype}
\usepackage{inconsolata}
\usepackage{graphicx}
\usepackage{float}

\newcommand{\inc}[1]{\textcolor{ForestGreen}{\ensuremath{_{\,\uparrow #1}}}}
\newcommand{\dec}[1]{\textcolor{BrickRed}{\ensuremath{_{\,\downarrow #1}}}}
\definecolor{lightgray}{gray}{0.95}

\title{StatABench: Dataset and Framework for Evaluating Statistical Analysis Capabilities of LLMs}

\author{
  Youxin Zhu$^{1}$\thanks{Equal contribution.},
  Yixuan Ding$^{4}$\footnotemark[1],
  Peng Lai$^{1}$\footnotemark[1], Longyue Wang$^2$, Bingyi Jing$^3$,
  \textbf{Guanhua Chen}$^{1}$\thanks{Corresponding author.} \\
  $^1$Southern University of Science and Technology,
  $^2$Alibaba Group \\
  $^3$The Chinese University of Hong Kong, Shenzhen,
  $^4$The University of Hong Kong \\
}

\begin{document}
\maketitle

\begin{abstract}
Statistical analysis is a broad, complex field requiring both domain knowledge and tool proficiency. While prior work has evaluated large language models (LLMs) in this domain, existing benchmarks remain limited in scope and format. To bridge this gap, we introduce StatABench (\textbf{Stat}istical \textbf{A}nalysis \textbf{Bench}mark), a benchmark designed to systematically assess LLMs' statistical analysis capabilities. StatABench comprises two complementary components: \textbf{Stat-Closed}, containing 404 questions across 18 statistical topics in multiple formats (multiple-choice, fill-in-the-blank, decision-making, and practical application), and \textbf{Stat-Open}, featuring 30 complex open-ended modeling tasks adapted from professional competitions. We evaluate diverse LLMs using the LangChain MCP framework and multiple data science agents, and assess Stat-Open solutions via a validated LLM-as-Judge protocol. Experiments show that even GPT-5.1 achieves only 68.6\% on Stat-Closed, while the best open-source model reaches 60.6\%. On Stat-Open, the top agent framework scores 61.86 on average. These results reveal the gap between current LLMs and reliable statistical analysis, highlighting persistent challenges in tool-grounded reasoning, methodological decision-making, and end-to-end statistical modeling. The code is available at \url{https://github.com/youxin01/StatABench}.
\end{abstract}

\begin{figure*}[t]
    \centering
    \includegraphics[width=0.9\textwidth]{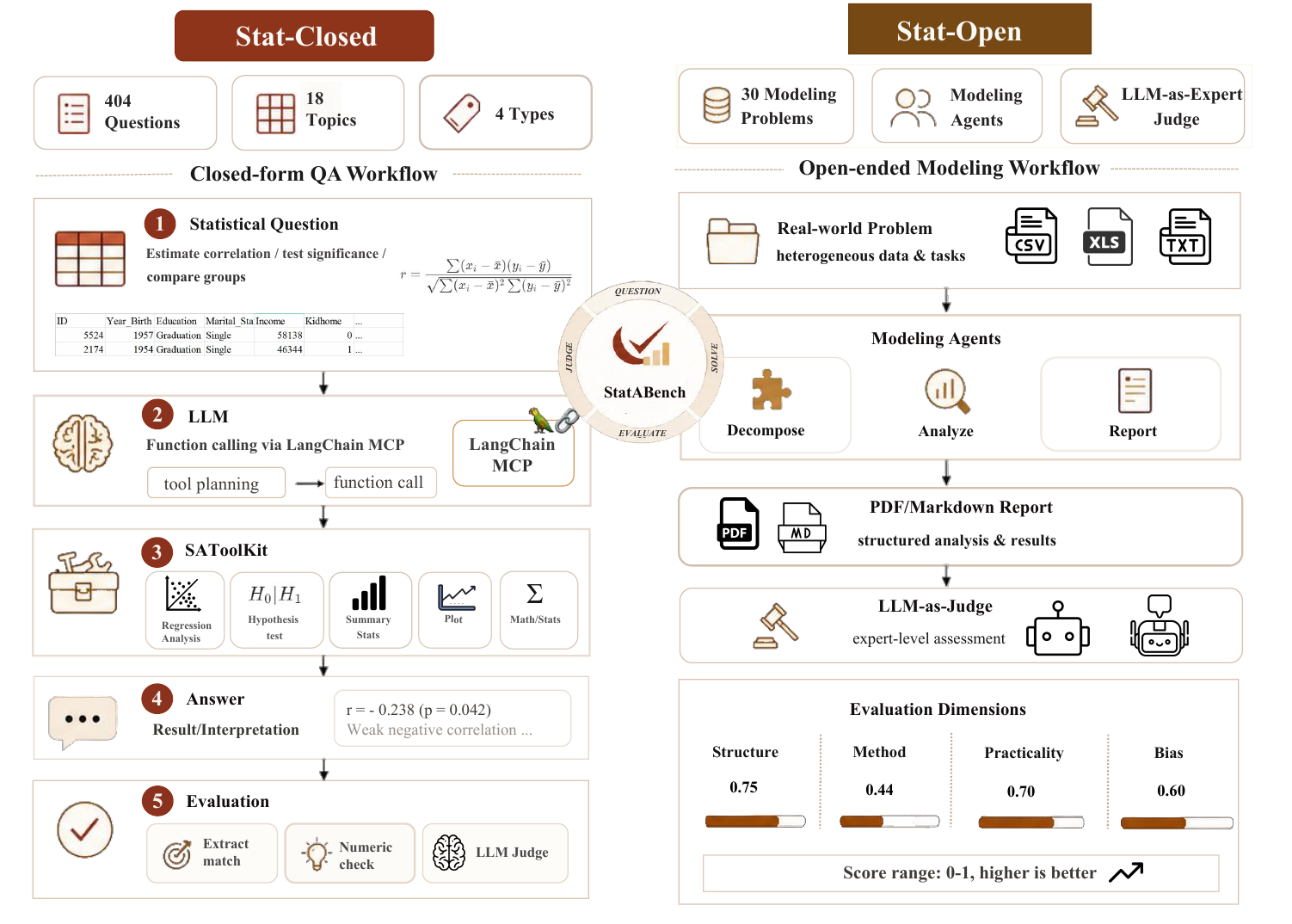}
    \caption{
        \textbf{Overview of StatABench.}
        StatABench consists of two complementary tracks for evaluating LLMs' statistical analysis capabilities.
        \textbf{Stat-Closed} includes 404 closed-form questions covering 18 statistical topics and 4 question types, where models solve statistical QA tasks with tool support and are evaluated by ground-truth-based matching.
        \textbf{Stat-Open} includes 30 open-ended modeling problems, where modeling agents conduct real-world statistical analysis and generate structured reports evaluated by an LLM-as-Judge system.
        }
\end{figure*}

\begin{table}[ht]
    \centering
    \resizebox{\columnwidth}{!}{
        \begin{tabular}{lcccc}
        \toprule
        \textbf{Benchmark} & \textbf{Tool} & \textbf{Open-} & \textbf{Real-} & \textbf{Eval.} \\
        & \textbf{Use} & \textbf{Ended} & \textbf{World} & \textbf{Type} \\
        \midrule
        StatQA \cite{zhu2024statqa} & \xmark & \xmark & \cmark & Static \\
        QRData \cite{liu2024llmscapabledatabasedstatistical} & \xmark & \xmark & \cmark & Static \\
        StatLLM \cite{song2025statllmdatasetevaluatingperformance} & \cmark\textsuperscript{*} & \xmark & \xmark & Static \\
        StatEval \cite{lu2025statevalcomprehensivebenchmarklarge} & \xmark & \xmark & \xmark & Static \\
        DSBench \cite{jing2025dsbench} & \cmark & \xmark & \cmark & Static \\
        DataSciBench \cite{li2025datascibench} & \cmark & \xmark & \cmark & Static \\
        \midrule
        \textbf{StatABench (Ours)} & \cmark & \cmark & \cmark & \textbf{Dynamic} \\
        \bottomrule
        \multicolumn{5}{l}{\footnotesize \textsuperscript{*}Limited to rigid code generation (e.g., SAS).}
        \end{tabular}}
        \caption{Comparison with existing benchmarks. StatABench uniquely combines statistical domain focus with agentic tool use, open-ended modeling, and dynamic LLM-as-Judge evaluation.}
        \label{tab:comparison}
    \end{table}
    
\section{Introduction}
Across healthcare, finance, and the social sciences, statistical analysis provides the methodological foundation for drawing conclusions from data and supporting evidence-based decisions \citep{moore1991statistics}. With the rapid advancement of large language models (LLMs) \citep{qin2025largelanguagemodelsmeet}, growing interest has emerged in leveraging them for statistical analysis tasks \citep{zhu2024statqa,liu2024llmscapabledatabasedstatistical,ji2025overviewlargelanguagemodels}. This potential is particularly important because real-world statistical analysis often begins with underspecified natural-language questions but ultimately requires executable procedures, appropriate tool use, and context-aware interpretation.

A key insight motivating our work is that \textit{real-world statistical analysis is inseparable from tool use}. A statistician does not merely reason about which test to apply---they must also correctly invoke software (R, Python, SPSS) with precise parameters and interpret the output. We therefore argue that evaluating statistical analysis capability requires assessing both conceptual understanding and the ability to operationalize that understanding through tools. This dual requirement distinguishes our benchmark from prior work that evaluates either pure reasoning or general tool use in isolation.

However, existing benchmarks \citep{song2025statllmdatasetevaluatingperformance, zhu2024statqa, liu2024llmscapabledatabasedstatistical} generally suffer from narrow domain scope, insufficient task difficulty, or rigid format constraints. To address these limitations, we propose StatABench, comprising two complementary components:

\textbf{Stat-Closed} contains 404 problems spanning 18 major statistical topics and multiple question types: multiple-choice, fill-in-the-blank, decision-making, and practical application. As the baseline evaluation setting, we leverage the LangChain MCP Framework \citep{langchain_mcp} to integrate models with our Statistical Analysis Toolkit (SAToolKit), which provides 35 callable statistical analysis functions. We further evaluate performance using four advanced data science agents---Qwen Agent \citep{qwen_agent}, CrewAI \citep{crewai}, AutoGen \citep{autogen}, and Smolagents \citep{smolagents}---to assess LLM performance under different agentic workflows.

\textbf{Stat-Open} targets complex, open-ended, real-world statistical analysis. We curate 30 challenging problems from renowned modeling competitions (MCM/ICM, CUMCM, MAS), where each task undergoes rigorous verification by human domain experts to ensure statistical relevance and solvability. Unlike Stat-Closed, these tasks require comprehensive problem understanding, advanced tool integration, and deep reasoning. We employ different modeling frameworks \citep{liu2025mmagent,qian2025modelingagent} that output structured modeling reports, evaluated via a validated LLM-as-Judge protocol with demonstrated human-machine alignment.

Our experiments reveal that even GPT-5.1 achieves only 68.6\% accuracy on Stat-Closed, while the best open-source model, Qwen2.5-72B, reaches 60.6\%. On Stat-Open, the top agent framework achieves an average score of 61.86. Together with statistical significance tests ($p = 2.87 \times 10^{-9}$, Friedman test), these results indicate substantial room for improvement before LLMs can serve as reliable statistical analysis experts.

\section{Related Work}
\paragraph{Reasoning and statistical benchmarks.}
Early LLM benchmarks focused on knowledge recall and common-sense reasoning \citep{hendrycks2021measuringmassivemultitasklanguage,srivastava2023imitationgamequantifyingextrapolating}, and were later extended to mathematical reasoning (GSM8K \citep{cobbe2021trainingverifierssolvemath}, UGMathBench \citep{xu2025ugmathbenchdiversedynamicbenchmark}) and code generation (HumanEval \citep{chen2021evaluatinglargelanguagemodels}, MBPP \citep{austin2021programsynthesislargelanguage}). While these benchmarks cover diverse reasoning abilities, they largely assume well-defined problems with single correct answers and therefore do not reflect the interpretive and tool-dependent nature of statistical analysis.

Statistics-focused benchmarks partially address this gap. StatQA \citep{zhu2024statqa} covers five topics with template-generated questions; QRData \citep{liu2024llmscapabledatabasedstatistical} focuses on quantitative reasoning over structured data; StatLLM \citep{song2025statllmdatasetevaluatingperformance} broadens topical coverage but is tied to SAS code generation; and StatEval \citep{lu2025statevalcomprehensivebenchmarklarge}, despite its $\sim$20K problems, remains text-only and primarily probes statistical reasoning. Consequently, existing statistical benchmarks still fall short of evaluating authentic statistical analysis, where models must handle data, select appropriate methods, use tools, and interpret results in context.

\paragraph{Data-science agent benchmarks.}
Recent work has begun evaluating LLMs as autonomous data-science agents. DSBench \citep{jing2025dsbench} introduces 466 Kaggle-style analysis tasks; DataSciBench \citep{li2025datascibench} provides semi-automated ground truth across diverse data-science tasks; and DSAEval \citep{sun2026dsaevalevaluatingdatascience} scales to 641 problems across structured and unstructured data. These benchmarks move closer to realistic tool-based workflows, but they primarily evaluate \textit{generic} data-science capabilities rather than the rigorous application of statistical methodology, such as hypothesis testing, regression diagnostics, Bayesian inference, and causal estimation. In contrast, StatABench focuses specifically on statistical analysis while retaining the tool-use and workflow complexity needed for practical evaluation.

\paragraph{Evaluation methodologies.}
LLM evaluation is shifting from static QA to dynamic and agentic scenarios. Toolformer \citep{schick2023toolformerlanguagemodelsteach} and ReAct \citep{yao2023reactsynergizingreasoningacting} demonstrated tool-augmented reasoning, motivating evaluations that consider not only final answers but also intermediate tool use. For open-ended assessment, the LLM-as-a-Judge paradigm \citep{ashktorab2025evalassisthumancenteredtoolllmasajudge,zheng2023judgingllmasajudgemtbenchchatbot} has become widely adopted, with recent work exploring its extension to Agent-as-a-Judge for complex agentic tasks \citep{zhuge2025agentjudge}. Closely related to our setting, \citet{qian2025modelingagent} employed LLMs to evaluate mathematical modeling reports from multiple perspectives, supporting the applicability of judge-based evaluation for structured analytical outputs. Building on these developments, StatABench uses LLM-as-Judge to evaluate open-ended statistical modeling reports while validating its alignment with human judgments.

\section{The StatABench Benchmark}
\label{sec:construct_data}
\subsection{Overview}

\textbf{Stat-Closed} comprises 404 problems with ground truths across multiple formats: multiple-choice, fill-in-the-blank, decision-making, and practical application. Practical application tasks are accompanied by datasets, requiring models to analyze data and derive conclusions using our SAToolKit. The 18 statistical topics (Table~\ref{tab:task_counts}) span from foundational areas like descriptive statistics and confidence intervals to advanced topics including causal inference, survival analysis, and high-dimensional data analysis.

\begin{table}[t]
\centering
\resizebox{0.9\columnwidth}{!}{
\begin{tabular}{lc}
    \toprule
    \textbf{Task Category (Abbr.)} & \textbf{Count} \\
    \midrule
    \rowcolor{lightgray} Nonparametric Statistics (NS) & 30 \\
    Descriptive Statistics (DS) & 30 \\
    \rowcolor{lightgray} Multiple Comparison (MC) & 30 \\
    Correlation Analysis (CA) & 30 \\
    \rowcolor{lightgray} Causal Inference (CausalI) & 30 \\
    Distribution Compliance Test (DCT) & 30 \\
    \rowcolor{lightgray} Generalized Linear Model (GLM) & 25 \\
    Survival Analysis (SA) & 20 \\
    \rowcolor{lightgray} Point Estimation (PE) & 20 \\
    High Dimensional Data (HDD) & 20 \\
    \rowcolor{lightgray} Contingency Table Test (CTT) & 20 \\
    AB Test (AB) & 20 \\
    \rowcolor{lightgray} Regression (Reg) & 20 \\
    Variance Test (VT) & 20 \\
    \rowcolor{lightgray} Bayesian Method (BM) & 18 \\
    Time Series (TS) & 15 \\
    \rowcolor{lightgray} Exploratory Data Analysis (EDA) & 15 \\
    Confidence Interval (CI) & 11 \\
    \midrule
    \textbf{Total} & \textbf{404} \\
    \bottomrule
    \end{tabular}}
    \caption{\textbf{Statistics of Stat-Closed.} Distribution of 404 questions across 18 categories.}
    \label{tab:task_counts}
\end{table}

\textbf{Stat-Open} targets complex, open-ended reasoning where no standard answer exists, containing 30 tasks adapted from mathematical modeling contests. These problems are designed to evaluate LLMs in three core capabilities: solving complex reasoning tasks, comprehending statistical concepts, and generating structured analytical reports. Each task requires the agent to perform end-to-end data cleaning, exploratory analysis, model building, and report generation.

\subsection{Data Collection for Stat-Closed}
\begin{figure*}[!htbp]
    \centering
    \includegraphics[width=0.85\linewidth]{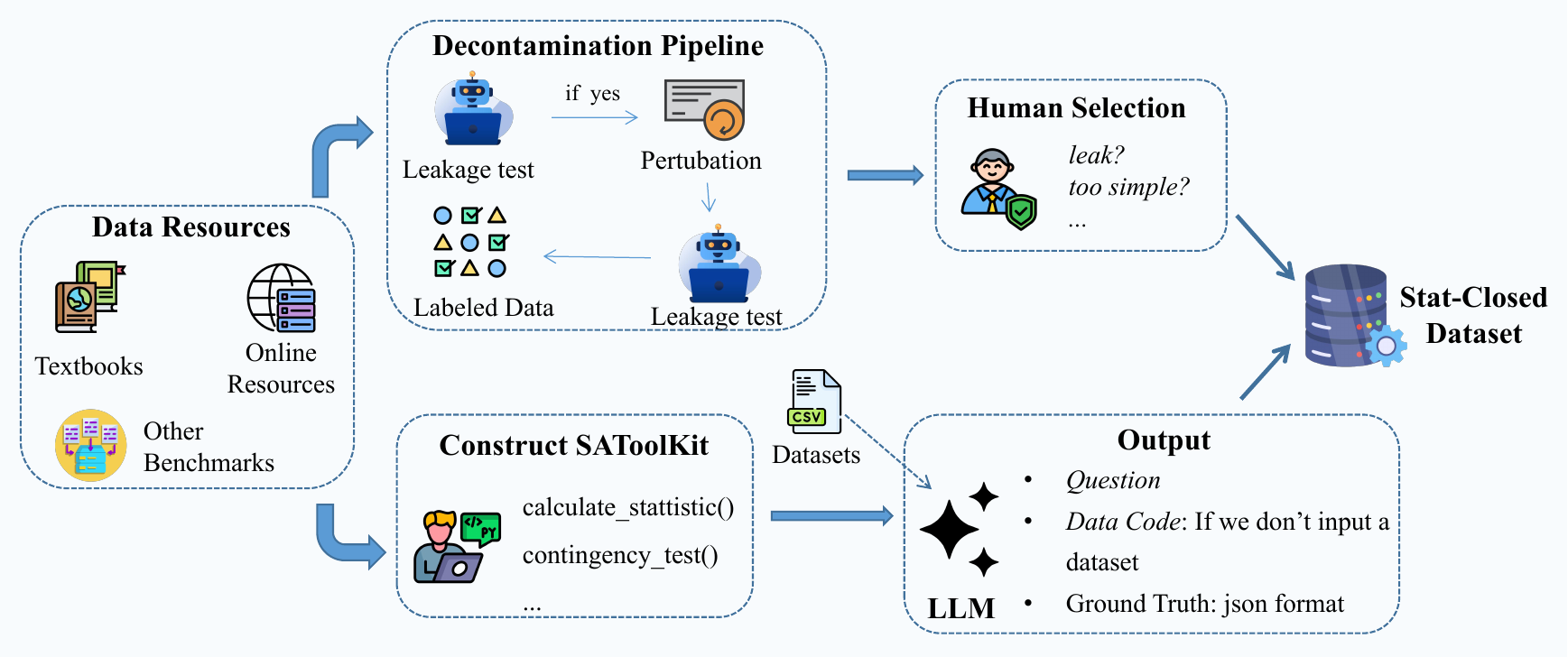}
    \caption{The data collection pipeline for Stat-Closed. \textbf{Top:} Construction of standard statistical questions with a \textit{Modify-and-Verify} decontamination pipeline. \textbf{Bottom:} Generation of practical application tasks via two pathways---adapting existing datasets or generating synthetic data.}
    \label{fig:buid_close}
\end{figure*}

We curate Stat-Closed from 18 major statistical topics, ensuring diversity in domain coverage and difficulty. The process can be divided into 4 parts (Figure \ref{fig:buid_close}).

\paragraph{Source Aggregation.} We aggregated questions from three complementary sources: (1) open-sourced statistical textbooks covering probability, statistical learning, and Bayesian analysis; (2) existing benchmarks \citep{zhu2024statqa, jin2024cladderassessingcausalreasoning, jin2024largelanguagemodelsinfer}; and (3) public online resources including open-access repositories and examination sets. This multi-source strategy ensures broad coverage of both theoretical foundations and applied scenarios. Details are provided in Appendix~\ref{sec:source_details}.

\paragraph{Decontamination Pipeline.} To mitigate data contamination risk \citep{xu2025dcrquantifyingdatacontamination, chen2025recentadvanceslargelangauge}, we implemented a strict modify-and-verify pipeline (see Algorithm~\ref{alg:decontamination}). Questions correctly answered by Qwen2.5-72B-Instruct\citep{qwen2025qwen25technicalreport} and Llama-3.1-70B-Instruct\citep{llama3.1} were flagged as potential leakage and manually perturbed---including rewriting problem stems, adding distractors, shuffling options, or inverting True/False conditions. Only instances where at least one model subsequently failed, or those explicitly verified as high-quality by domain experts, were retained. This procedure reduces the risk of data leakage by making it less likely that models can solve the benchmark through memorized public examples rather than genuine statistical reasoning.

\paragraph{SAToolKit.}
To support tool-based evaluation in Stat-Closed, we implement the Statistical Analysis Toolkit (SAToolKit), a collection of 35 callable statistical functions covering all 18 topics. SAToolKit provides a unified interface for common statistical procedures, ranging from descriptive statistics to inferential analysis. By exposing the same function set to all models, SAToolKit allows us to evaluate whether LLMs can select appropriate statistical tools, construct valid parameters, and interpret analytical outputs in a controlled and reproducible setting.

\paragraph{Practical Application Tasks.}
Based on SAToolKit, we construct practical application questions as triplets consisting of a statistical question, an associated dataset, and ground-truth function parameters. We employed Gemini 2.5 Pro \citep{comanici2025gemini25pushingfrontier} to generate these instances through two complementary pathways:

\begin{itemize}[nosep,leftmargin=*]
    \item \textit{Dataset-Driven Adaptation:} We first identify a problem type (e.g., normality test) and its corresponding SAToolKit function, then select a high-quality dataset suited to this type. By prompting Gemini 2.5 Pro with both the dataset content and the function definition, we guide it to formulate a relevant statistical question that requires analyzing the data and inferring the precise function parameters.
    \item \textit{Synthetic Generation:} When suitable public datasets are unavailable, we provide the model with the target SAToolKit function and prompt it to generate a corresponding statistical question, Python code for dataset synthesis, and the correct function parameters. The generated code is then executed to obtain the dataset.
\end{itemize}

All generated instances were manually reviewed by authors with statistical backgrounds. We checked the clarity of each question, the consistency between the dataset and the intended statistical concept, the correctness of the selected SAToolKit function and its ground-truth parameters, and the reproducibility of the reference answer. Ambiguous or inconsistent instances were revised or discarded.

\subsection{Data Collection for Stat-Open}

Stat-Open evaluates LLMs on \textit{complex, open-ended, real-world statistical analysis}. These tasks require not only the application of statistical techniques but also comprehensive problem understanding, multi-step reasoning, and integration of domain knowledge.

We curated 30 problems from multiple years of three renowned competitions: MCM/ICM (Mathematical Contest in Modeling / International Mathematical Contest in Modeling), CUMCM (China Undergraduate Mathematical Contest in Modeling), and MAS (National Case Competition for Applied Statistics Postgraduates).. Each problem incorporates realistic scenario settings, poses questions related to statistical modeling, and is accompanied by corresponding data files. Unlike prior mathematical modeling benchmarks, these problems impose higher requirements on specialized statistical knowledge and mandate analysis from a data science perspective, while outputting structured analytical documents to facilitate evaluation.

Each Stat-Open task is not a single isolated question, but a complete statistical modeling project. Solving one task typically requires decomposing the problem into multiple stages, including problem understanding, data cleaning, method selection, code execution, result interpretation, and report writing. Therefore, the difficulty and discriminative power of Stat-Open arise not only from the number of tasks, but also from the depth and breadth of the workflow that each task entails.

\begin{figure}[ht]
    \centering
    \includegraphics[width=\columnwidth]{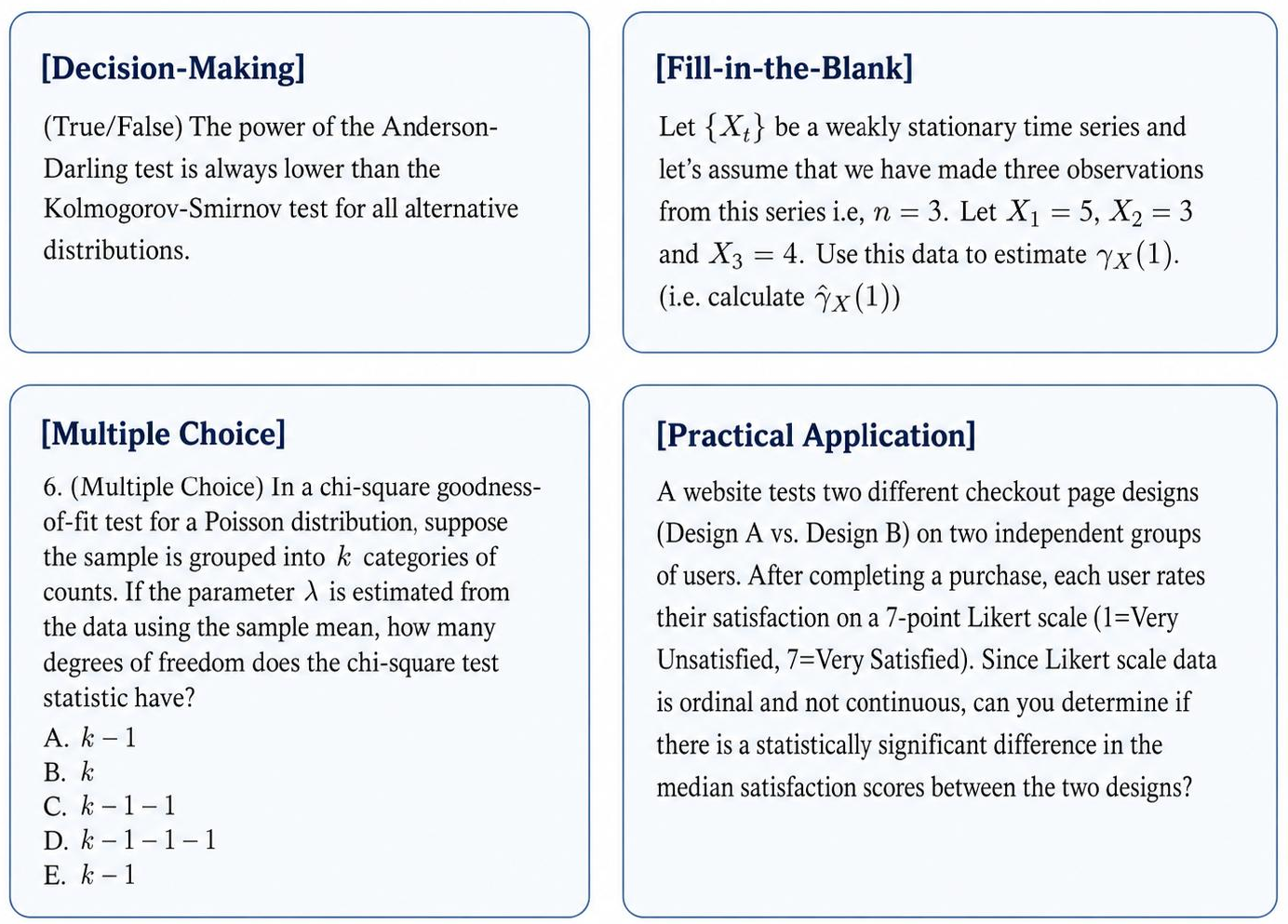}
    \caption{Example questions in Stat-Closed illustrating the four types: Decision-Making, Fill-in-the-blank, Multiple-Choice, and Practical Application.}
    \label{fig:ex_ques}
\end{figure}

\subsection{Annotations and Evaluation}
\paragraph{Stat-Closed.}
Most questions come with well-defined ground truths from their original sources. For practical application tasks, we first execute the corresponding Python function with the correct parameters to obtain the deterministic result, and then provide both the result and the original question to DeepSeek-V3 \citep{deepseekai2025deepseekv3technicalreport} to generate a human-readable final answer as the reference output.

For evaluation, we extract the final answer from each model response and compare it against the reference output. We use direct matching for multiple-choice, decision-making, and numerical questions. For open-form responses, such as fill-in-the-blank questions with semantically equivalent expressions and practical application questions requiring textual interpretation, we adopt the LLM-as-Judge paradigm \citep{ashktorab2025evalassisthumancenteredtoolllmasajudge}. The judge is given the original question, the reference answer, and the model prediction, and determines whether the prediction is consistent with the reference in terms of statistical conclusion, numerical result, and interpretation.

\paragraph{Stat-Open.}
\label{eval:hard}
Given the open-ended nature of these tasks, we adopt the \textbf{LLM-as-Judge} paradigm. Our evaluation rubric is adapted from validated frameworks in mathematical modeling evaluation \citep{qian2025modelingagent,liu2025mmagent}, which have demonstrated that multi-dimensional assessment can effectively capture the quality of complex analytical outputs. We assess reports across four dimensions, each targeting a distinct aspect of statistical modeling competence:
\vspace{-4pt}
\begin{itemize}
    \setlength{\itemsep}{2pt}    
    \setlength{\parsep}{2pt}     
    \setlength{\parskip}{2pt}  
    \item \textbf{Structural Coherency:} Assesses the report's structural integrity and technical depth by assessing the rigor and completeness of all essential modeling components 
    \item \textbf{Methodological Quality:} Examines the methodology from four key perspectives: the rigor of modeling techniques, the authenticity of data application, the depth of analytical reasoning, and the innovativeness of the approach.
    \item \textbf{Practicality and Scientific Validity:} Evaluates the robustness and the practical validity of the proposed model.
    \item \textbf{Result Interpretation and Bias Analysis:} Evaluates the report's ability to identify and mitigate potential biases in data or modeling, ensuring the conclusions are trustworthy and defensible.
\end{itemize}
\vspace{-2pt}
These dimensions collectively cover the full lifecycle of statistical modeling---from problem formulation through methodology to interpretation---ensuring that evaluation captures both technical rigor and practical utility. Final scores are computed as an equally weighted average across all dimensions, following \citet{qian2025modelingagent}.

\section{Experiments}
\label{sec:experiments}

\subsection{Stat-Closed}
\label{sec:exp_easy}
\begin{table*}[t]
    \centering
    \caption{\textbf{Task-specific accuracy (\%) on Stat-Closed.} \textbf{Bold} = best per column. Gray column = overall accuracy. Models above the line are open-source; below are closed-source.}
    \label{mcp_res}
    \resizebox{\textwidth}{!}{
    \begin{tabular}{l cccccccccccccccccc >{\columncolor{lightgray}}c}
    \toprule
    \textbf{Model} & \textbf{AB} & \textbf{BM} & \textbf{CA} & \textbf{CI} & \textbf{CTT} & \textbf{CausalI} & \textbf{DCT} & \textbf{DS} & \textbf{EDA} & \textbf{GLM} & \textbf{HDD} & \textbf{MC} & \textbf{NS} & \textbf{PE} & \textbf{Reg} & \textbf{SA} & \textbf{TS} & \textbf{VT} & \textbf{Total} \\
    \midrule
    Llama3.1-8B & 35.0 & 33.3 & 13.3 & 18.2 & 5.0 & 40.0 & 16.7 & 33.3 & 46.7 & 8.0 & 40.0 & 23.3 & 43.3 & 5.0 & 35.0 & 10.0 & 6.7 & 40.0 & 25.5 \\
    Qwen3-8B & 45.0 & 50.0 & 30.0 & 18.2 & 10.0 & 53.3 & 33.3 & \textbf{76.7} & 53.3 & 20.0 & 45.0 & 46.7 & 26.7 & 50.0 & 50.0 & 35.0 & 20.0 & 30.0 & 39.6 \\
    Qwen2.5-7B & 50.0 & 27.8 & 23.3 & 18.2 & 25.0 & 66.7 & 26.7 & 40.0 & 66.7 & 32.0 & 35.0 & 46.7 & 60.0 & 25.0 & 50.0 & 30.0 & 26.7 & 15.0 & 38.1 \\
    Qwen2.5-32B & 40.0 & 66.7 & 53.3 & 27.3 & 15.0 & 63.3 & 36.7 & \textbf{76.7} & 73.3 & 40.0 & \textbf{60.0} & 56.7 & 40.0 & 70.0 & 60.0 & 30.0 & 53.3 & 60.0 & 51.7 \\
    Qwen2.5-72B & 50.0 & 66.7 & 70.0 & 45.5 & 40.0 & 73.3 & 46.7 & 73.3 & \textbf{80.0} & 32.0 & 35.0 & 66.7 & 73.3 & 70.0 & 65.0 & 50.0 & \textbf{80.0} & \textbf{65.0} & 60.6 \\
    GPT-4o-mini & 35.0 & 55.6 & 40.0 & 9.1 & 0.0 & 73.3 & 10.0 & 73.3 & 66.7 & 36.0 & 30.0 & 66.7 & 46.7 & 60.0 & 35.0 & 25.0 & 53.3 & 35.0 & 43.3 \\
    DeepSeek-V3 & \textbf{75.0} & 44.4 & 53.3 & 63.6 & 70.0 & 73.3 & 56.7 & \textbf{76.7} & 60.0 & 52.0 & 55.0 & 53.3 & 53.3 & 40.0 & 75.0 & 30.0 & 60.0 & 35.0 & 57.4 \\
    GPT-5.1 & 55.0 & \textbf{66.7} & \textbf{76.7} & 54.5 & 65.0 & 73.3 & 56.7 & \textbf{83.3} & 73.3 & \textbf{68.0} & \textbf{85.0} & \textbf{70.0} & 73.3 & 55.0 & 70.0 & \textbf{80.0} & 66.7 & 45.0 & \textbf{68.6} \\
    Claude~4.5 & 55.0 & 50.0 & 63.3 & 45.5 & \textbf{80.0} & \textbf{73.3} & 33.3 & \textbf{86.7} & 66.7 & 72.0 & 60.0 & \textbf{70.0} & \textbf{86.7} & \textbf{70.0} & \textbf{75.0} & \textbf{80.0} & 60.0 & 45.0 & 66.3 \\
    \bottomrule
    \end{tabular}
    }
\end{table*}

\begin{table*}[t]
    \centering
    \caption{\textbf{Accuracy (\%) on practical application tasks with different agent frameworks.} LangChain MCP serves as the baseline (second column). Subscripts show absolute improvement in percentage points. DeepSeek-V3 + CrewAI reaching 85.35\% confirms questions are well-defined, while most combinations cluster in 50--60\%.}
    \resizebox{0.9\textwidth}{!}{
    \begin{tabular}{lccccc}
        \toprule
        \textbf{Model} & \textbf{LangChain MCP} & \textbf{Qwen Agent} & \textbf{AutoGen} & \textbf{CrewAI} & \textbf{Smolagents} \\
        \midrule
        Qwen2.5-7B & 23.74 & 24.75\inc{1.01} & 40.40\inc{16.66} & 41.92\inc{18.18} & 24.75\inc{1.01} \\
        Qwen2.5-32B & 41.41 & 53.03\inc{11.62} & 43.43\inc{2.02} & 70.71\inc{29.30} & 57.07\inc{15.66} \\
        Qwen2.5-72B & \textbf{57.07} & 59.60\inc{2.53} & 68.18\inc{11.11} & 76.26\inc{19.19} & 52.53\dec{4.54} \\
        GPT-4o-mini & 25.76 & 46.97\inc{21.21} & 41.92\inc{16.16} & 68.18\inc{42.42} & 38.38\inc{12.62} \\
        DeepSeek-V3 & 53.54 & \textbf{66.67}\inc{13.13} & \textbf{80.30}\inc{26.76} & \textbf{85.35}\inc{31.81} & \textbf{74.75}\inc{21.21} \\
        \bottomrule
    \end{tabular}}
  \label{agent_res}
\end{table*}

\paragraph{Setup.} 
We integrate SAToolKit with target models through LangChain MCP as the baseline tool-use setting. We evaluate Qwen3-8B \citep{yang2025qwen3technicalreport}, Qwen2.5-Instruct (7B, 32B, 72B) \citep{qwen2025qwen25technicalreport}, Llama-3.1-8B-Instruct \citep{llama3.1}, DeepSeek-V3 \citep{deepseekai2025deepseekv3technicalreport}, GPT-4o-mini \citep{gpt4o_mini}, GPT-5.1 \citep{gpt51}, and Claude-Sonnet-4.5-20250929 (Claude 4.5) \citep{claude_sonnet_45}. To ensure deterministic and reproducible evaluation, all models are run with greedy decoding, i.e., temperature 0. We additionally evaluate four advanced agent frameworks---Qwen Agent, CrewAI, AutoGen, and Smolagents---on the 198 practical application questions to test Stat-Closed under stronger agent-based settings. Details are provided in Appendix~\ref{app:exp_easy}.

\subsubsection{Results and Analysis}

\begin{figure}[ht]
\centering
  \includegraphics[width=0.95\columnwidth]{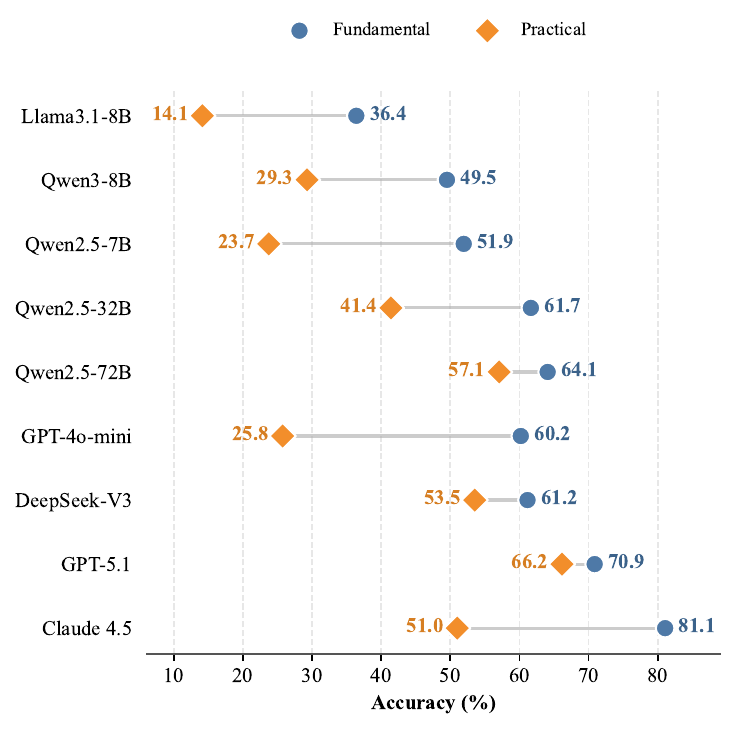}
  \caption{\textbf{The tool-integration tax.} Blue markers = fundamental statistical tasks; orange = practical application tasks requiring tool use.}
  \label{fig:too_nontool}
\end{figure}

\paragraph{Stat-Closed presents significant challenges.}
As shown in Table~\ref{mcp_res}, even GPT-5.1 reaches only 68.6\% overall accuracy, while Qwen2.5-72B, the best-performing open-source model among those evaluated, reaches 60.6\%. A Friedman test across 18 task categories yields $p = 2.87 \times 10^{-9}$, confirming that the model differences are statistically significant rather than due to random variation. Figure~\ref{fig:too_nontool} reveals a consistent tool-integration tax: all models perform worse on practical SAToolKit-based questions than on fundamental statistical questions. This drop is especially pronounced for smaller or less tool-robust models, such as Llama3.1-8B and GPT-4o-mini, while stronger models such as GPT-5.1 show a smaller but still visible decrease. Notably, even Claude 4.5, which achieves the highest score on fundamental questions, suffers a substantial drop in the practical setting. These results indicate that practical statistical analysis requires more than conceptual understanding: models must also translate statistical intent into correct tool choices, parameter settings, and result interpretation. This gap is precisely the dual capability that StatABench is designed to evaluate.

\paragraph{Model differences mainly stem from statistical errors.}
Figure~\ref{fig:too_nontool} shows that all models incur a tool-integration tax, but the magnitude of this penalty varies substantially across models even under the same LangChain MCP setup (Table~\ref{mcp_res}). To understand the source of this variation, we analyze failures using the taxonomy in Table~\ref{tab:error_analysis}, which separates statistical errors from engineering errors. Statistical errors include inappropriate tool selection (e.g., choosing \texttt{detect\_outliers} when the task requires \texttt{mood\_variance\_test}) and incorrect result interpretation (e.g., misreading a p-value), whereas engineering errors include format violations and parameter type mismatches. Among the 38 cases in which DeepSeek-V3 fails but Qwen2.5-72B succeeds, 74\% are statistical errors and 26\% are engineering errors. This distribution suggests that the observed cross-model gap is driven primarily by differences in statistical capability rather than framework-specific implementation artifacts.

\paragraph{Framework choice rescales but does not erase the gap.}
Table~\ref{agent_res} shows that stronger agent frameworks substantially improve absolute performance on Stat-Closed. For example, CrewAI raises DeepSeek-V3 to 85.35\%, indicating that the benchmark is well-posed and that many tasks become solvable under stronger orchestration. However, these improvements do not eliminate cross-model stratification: most framework settings still place model performance in the 50--60\% range. The gains are also highly model-dependent, with GPT-4o-mini improving by 42.4 percentage points under CrewAI, DeepSeek-V3 by 31.8 points, and Qwen2.5-72B by 19.2 points. This pattern suggests that stronger frameworks primarily alleviate execution and coordination bottlenecks, while the underlying statistical capability gap across models remains. The same tendency appears on Stat-Open, where even DeepSeek-V3 paired with the strongest framework achieves an average score of only 61.86 out of 100 (Table~\ref{tab:hard2}).

\subsection{Stat-Open}
\label{sec:exp_hard}

We evaluate Stat-Open with MathModelAgent\footnote{\url{https://github.com/jihe520/MathModelAgent}} and LLM-MM-Agent \citep{liu2025mmagent}, two agent frameworks for end-to-end statistical analysis through tool use, problem decomposition, and report synthesis. To isolate framework effects, we instantiate both agents with the same backbone model, DeepSeek-V3. The generated reports are then scored by Gemini 3 Pro \citep{gemini3pro} using the dimensions defined in \S\ref{sec:construct_data}. We select Gemini 3 Pro as the judge because, on a subset annotated by four human raters, it achieves the highest average Fleiss' Kappa ($\bar{\kappa}=0.51$, versus $0.25$ for GPT-5 and $0.19$ for Grok-4) and leads on six of seven sub-criteria. We additionally report Fleiss' Kappa between the LLM judge and human annotators to quantify evaluation reliability; per-dimension agreement is provided in Appendix~\ref{app:judge_selection}, and calculation details are given in Appendix~\ref{sec:Fleiss_Kappa_Calculation}.

\begin{table*}[t]
\centering
\caption{\textbf{Results and Inter-Annotator Agreement on Stat-Open.} Scores are scaled to 0--100. Fleiss' Kappa quantifies agreement between LLM judge and human raters.}
\label{tab:hard2}
\resizebox{0.85\textwidth}{!}{
\begin{tabular}{lcccc}
    \toprule
    \multirow{2}{*}{\textbf{Metric}}
    & \multicolumn{2}{c}{\textbf{MathModelAgent}}
    & \multicolumn{2}{c}{\textbf{LLM-MM-Agent}} \\
    \cmidrule(lr){2-3} \cmidrule(lr){4-5}
    & \textbf{LLM / Human} & \textbf{Fleiss' $\kappa$}
    & \textbf{LLM / Human} & \textbf{Fleiss' $\kappa$} \\
    \midrule
    Structural Coherency   & 71.50 / 69.67 & 0.465  & 34.00 / 32.67 & 0.535 \\
    Modeling Groundedness  & 52.50 / 52.00 & 0.698  & 59.00 / 56.67 & 0.652 \\
    Data Groundedness      & 56.50 / 55.00 & 0.579  & 39.50 / 39.67 & 0.688 \\
    Analysis Groundedness  & 70.00 / 69.00 & 0.542  & 59.00 / 58.00 & 0.319 \\
    Innovativeness         & 14.50 / 20.67 & 0.228  & 65.50 / 63.33 & 0.359 \\
    Practical Science      & 91.00 / 87.17 & 0.092  & 69.00 / 66.33 & 0.520 \\
    Result Bias            & 77.00 / 74.67 & 0.284  & 54.00 / 51.67 & 0.489 \\
    \midrule
    \textbf{Average}       & 61.86 / 61.17 & 0.413  & 54.29 / 52.62 & 0.509 \\
    \bottomrule
\end{tabular}}
\end{table*}

\subsubsection{Results and Analysis}

\paragraph{Strengths in structure and practicality.}
Both frameworks achieve relatively high scores in Structural Coherency and Practical Science, suggesting that agent-based workflows can produce reports with recognizable structure and practical relevance. MathModelAgent performs particularly well in these dimensions (71.50 and 91.00 vs.\ 34.00 and 69.00), indicating stronger document structuring and practical-science alignment. This advantage is likely related to its more rigid template-based output design, which helps ensure that required sections are present. However, this rigidity may also limit flexibility, as reflected by MathModelAgent's lower Innovativeness score (14.50 vs.\ 65.50).

\paragraph{Structural completeness and innovativeness show a trade-off.}
Innovativeness appears more sensitive to agent design than other dimensions. MathModelAgent shows particularly low scores (14.50/20.67), while LLM-MM-Agent performs better (65.50/63.33), suggesting that its more flexible workflow may allow greater methodological diversity. This dimension, adapted from \citet{qian2025modelingagent}, also shows lower inter-annotator agreement ($\kappa \approx 0.2$--$0.4$), likely due to its inherent subjectivity. Nevertheless, the contrast between the two frameworks suggests an important trade-off in agent design: stronger structural constraints improve completeness, but may suppress methodological exploration.

\paragraph{Cross-rater consistency and evaluation reliability.}
LLM-MM-Agent achieves higher average Fleiss' Kappa (0.509 vs.\ 0.413), indicating stronger human-machine agreement across evaluation dimensions. This difference is especially clear in Data Groundedness ($\kappa = 0.688$ vs.\ $0.579$), suggesting that LLM-MM-Agent outputs are judged more consistently by human and LLM evaluators on data-related criteria. In contrast, MathModelAgent's more rigid report structure may align well with structural criteria but less consistently with subjective assessment dimensions.

To further validate evaluation reliability, we conducted a Wilcoxon Signed-Rank Test on per-dimension scores between two models of clearly different capacity (Qwen3-235B vs.\ Qwen2.5-VL-72B) on three sampled tasks. The test yields $p = 0.016$, suggesting that our evaluation protocol can capture capability differences across the seven dimensions. Per-dimension scores are reported in Appendix~\ref{app:wilcoxon}. Sensitivity analysis on discretization bins (8--16) shows stable dimension rankings (Appendix~\ref{app:sensitivity}).

\section{Conclusion}

We introduced StatABench, a benchmark for assessing LLMs' statistical analysis
capabilities through two components: Stat-Closed and Stat-Open. Stat-Closed
evaluates fundamental knowledge and tool use across 18 statistical topics with
404 expert-verified questions, while Stat-Open contains 30 open-ended modeling
tasks requiring end-to-end analysis, tool execution, and structured report
generation. Together, they cover statistical competence from closed-form problem
solving to realistic analytical workflows.

Across both components, our results reveal a substantial gap between current
LLMs and reliable statistical analysis. On Stat-Closed, even GPT-5.1 reaches only
68.6\%, and all models incur a tool-integration tax when translating statistical
concepts into SAToolKit operations. Stronger frameworks improve performance but
do not remove cross-model stratification. Error analysis shows that most
differences stem from statistical failures, such as inappropriate tool selection
and result interpretation. On Stat-Open, advanced agent frameworks achieve only
61.86 out of 100, indicating that end-to-end statistical modeling remains far
from saturation.

These findings suggest that progress on statistical analysis agents requires more
than stronger base models or orchestration. Future systems must connect
statistical reasoning with executable workflows, tool use, and defensible
interpretation across task settings. By combining closed-form tasks with
open-ended modeling problems, StatABench provides a rigorous testbed for
diagnosing weaknesses, distinguishing failure sources, and tracking progress in
statistical analysis with LLMs.

\section*{Limitations}

StatABench prioritizes task quality and diversity over raw scale. Although the current benchmark already supports meaningful discrimination among models, expanding both Stat-Closed and Stat-Open would enable more fine-grained evaluation across statistical topics and workflow stages.

In addition, benchmark maintenance remains an ongoing challenge. As models evolve and public tasks become more widely available, future updates will be needed to broaden model coverage and reduce potential contamination over time.

\section*{Ethics and Privacy Statement}

StatABench is constructed from open-access repositories, textbooks, and publicly available competition materials under their respective licenses (GPL-3.0, MIT, CC-BY-NC-SA). The dataset contains no personally identifiable information. Human evaluation was conducted by informed graduate-level volunteers. The benchmark is intended for research purposes only and should not replace professional statistical judgment in high-stakes settings.

\section*{Acknowledgements}
Large language models (Gemini 2.5 Pro and DeepSeek-V3) were employed in the data construction pipeline (Section~\ref{sec:construct_data}). The intellectual content, experimental design, and scientific conclusions remain exclusively those of the authors.

\bibliography{custom}

\appendix

\section{Judge Model Selection}
\label{app:judge_selection}

We compared three candidate judge models by computing Fleiss' Kappa between each model's ratings and human annotations across all evaluation dimensions (Table~\ref{tab:judge_compare}).

\begin{table}[h]
\centering
\caption{Fleiss' Kappa for candidate judge models.}
\label{tab:judge_compare}
\resizebox{\columnwidth}{!}{
\begin{tabular}{lccc}
    \toprule
    \textbf{Metric} & \textbf{Gemini 3 Pro} & \textbf{GPT-5} & \textbf{Grok-4} \\
    \midrule
    Structural Coherency & \textbf{0.535} & 0.049 & 0.209 \\
    Modeling Groundedness & \textbf{0.652} & 0.392 & 0.329 \\
    Data Groundedness & \textbf{0.688} & 0.241 & 0.245 \\
    Analysis Groundedness & \textbf{0.319} & 0.154 & 0.065 \\
    Innovativeness & 0.359 & \textbf{0.360} & 0.170 \\
    Practical Science & \textbf{0.520} & 0.187 & 0.201 \\
    Result Bias & \textbf{0.489} & 0.357 & 0.130 \\
    \midrule
    \textbf{Average} & \textbf{0.509} & 0.249 & 0.193 \\
    \bottomrule
\end{tabular}}
\end{table}

\section{Sensitivity Analysis}
\label{app:sensitivity}

We tested the sensitivity of Fleiss' Kappa to the number of discretization bins used for score quantization (Table~\ref{tab:sensitivity_bins}). The overall ranking pattern remains stable across all settings.

\begin{table}[h]
\centering
\caption{Fleiss' Kappa under different bin counts. Rankings in parentheses.}
\label{tab:sensitivity_bins}
\resizebox{\columnwidth}{!}{
\begin{tabular}{lccccccc}
    \toprule
    \textbf{Bins} & \textbf{SC} & \textbf{MG} & \textbf{DG} & \textbf{AG} & \textbf{Inn.} & \textbf{PS} & \textbf{RB} \\
    \midrule
    8  & .517(4) & .771(1) & .545(3) & .571(2) & .314(6) & .010(7) & .410(5) \\
    10 & .602(4) & .735(1) & .723(2) & .646(3) & .221(6) & .092(7) & .365(5) \\
    12 & .465(4) & .698(1) & .579(2) & .542(3) & .228(6) & .092(7) & .284(5) \\
    14 & .596(3) & .706(1) & .519(4) & .705(2) & .148(6) & .092(7) & .374(5) \\
    16 & .388(4) & .609(1) & .460(2) & .434(3) & .148(7) & .180(6) & .258(5) \\
    \bottomrule
\end{tabular}}
\end{table}

\section{Evaluation Reliability: Wilcoxon Test}
\label{app:wilcoxon}

To verify that the LLM judge can reliably differentiate models of different capacity, we sampled three tasks from Stat-Open and ran two backbones of clearly different capacity---Qwen3-235B (qwen3-235b-a22b-instruct-2507) and Qwen2.5-VL-72B (qwen2.5-vl-72b-instruct)---through the same agent framework. Average per-dimension scores assigned by Gemini 3 Pro are shown in Table~\ref{tab:wilcoxon}. A Wilcoxon Signed-Rank Test on the seven paired dimension scores yields $p = 0.016$, indicating that the score difference between the two models is statistically significant despite the small sample size, and supporting the reliability of our evaluation protocol.

\begin{table}[h]
\centering
\caption{Per-dimension average scores (in $[0,1]$) assigned by Gemini 3 Pro to two backbones of differing capacity, averaged over three sampled Stat-Open tasks. SC = Structural Coherency, MG = Modeling Groundedness, DG = Data Groundedness, AG = Analysis Groundedness, Inn = Innovativeness, PS = Practical Science, RB = Result Bias.}
\label{tab:wilcoxon}
\resizebox{\columnwidth}{!}{
\begin{tabular}{lccccccc}
    \toprule
    \textbf{Model} & \textbf{SC} & \textbf{MG} & \textbf{DG} & \textbf{AG} & \textbf{Inn} & \textbf{PS} & \textbf{RB} \\
    \midrule
    Qwen3-235B   & 0.517 & 0.467 & 0.500 & 0.500 & 0.250 & 0.833 & 0.850 \\
    Qwen2.5-VL-72B & 0.333 & 0.300 & 0.300 & 0.233 & 0.100 & 0.450 & 0.667 \\
    \bottomrule
\end{tabular}}
\end{table}

\section{Fleiss's Kappa Calculation}
\label{sec:Fleiss_Kappa_Calculation}
Fleiss's Kappa quantifies the inter-annotator agreement between AI and human raters for five evaluation metrics, adjusted for chance agreement. Calculations are performed independently for each metric, following these mathematical steps:

\noindent\textbf{1. Discretization}
Continuous scores $s \in [0,1]$ are discretized into $k=12$ equal-width categorical bins: $B_j = [(j-1)/12, j/12)$ for $j=1,\dots,11$, and $B_{12} = [11/12, 1)$. Each score $s$ is mapped to a category label $l(s) \in \{1,2,\dots,12\}$.

\noindent\textbf{2. Agreement Matrix}
For $N$ samples, construct an $N \times k$ matrix $\mathbf{N}$ where $N_{ij}$ is the number of raters (AI + Human) assigning category $j$ to sample $i$ ($N_{ij} \in \{0,1,2,3,4\}$ for 4 raters).

\noindent\textbf{3. Observed/Expected Agreement}
- Observed mean agreement:
\begin{equation}
     \bar{P} = \frac{1}{N} \sum_{i=1}^N \frac{1}{n(n-1)} \left( \sum_{j=1}^k N_{ij}^2 - n \right)
\end{equation}
  where $n=4$ (number of raters).
  
- Expected chance agreement:
\begin{equation}
      \bar{P}_e = \sum_{j=1}^k p_j^2, \quad p_j = \frac{1}{Nn} \sum_{i=1}^N N_{ij}
\end{equation}

\noindent\textbf{4. Final Kappa}
\begin{equation}
    \kappa = \frac{\bar{P} - \bar{P}_e}{1 - \bar{P}_e}
\end{equation}

Separate $\kappa$ values are computed for each of the five metrics to measure AI-human alignment at the granular level.

\section{Data Source Details}
\label{sec:source_details}
\label{sec:source_details}
\subsection{Stat-Closed}
To ensure the diversity and high quality of Stat-Closed, we curated questions from open-sourced textbooks, existing open-source benchmarks, and public online resources. 

\textbf{Copyright Statement:} All data collection was conducted strictly for non-commercial research purposes. For questions sourced from copyrighted textbooks, we performed manual selection and substantial reformatting (e.g., rewriting problem stems, standardizing notation) to ensure compliance with fair use principles. For open-source benchmarks and datasets, we strictly adhered to their respective licenses.

\SetAlgoNlRelativeSize{0}
\begin{algorithm}[t]
\small
\caption{Modify-and-Verify Decontamination}
\label{alg:decontamination}
    \SetKwInOut{Input}{Input}
    \SetKwInOut{Output}{Output}
    \SetKwFunction{Perturb}{Perturb}
    \SetKwFunction{Verify}{ExpertVerify}
    \SetKwFunction{Screen}{AllCorrect}
    \Input{Question pool $\mathcal{Q}_{raw}$; Models $\mathcal{M}$}
    \Output{Benchmark $\mathcal{Q}_{final}$}
    $\mathcal{Q}_{final} \leftarrow \emptyset$\;
    \ForEach{$q \in \mathcal{Q}_{raw}$}{
        \uIf{\Screen{$q, \mathcal{M}$}}{
            $q' \leftarrow \Perturb{q}$\;
            \If{\textbf{not} \Screen{$q', \mathcal{M}$} \textbf{or} \Verify{$q'$}}{
                $\mathcal{Q}_{final} \leftarrow \mathcal{Q}_{final} \cup \{q'\}$\;
            }
        }
        \Else{
            $\mathcal{Q}_{final} \leftarrow \mathcal{Q}_{final} \cup \{q\}$\;
        }
    }
    \Return{$\mathcal{Q}_{final}$}
\end{algorithm}

\paragraph{Textbooks}
We selected foundational exercises from the following open-access textbooks. These questions were adapted to fit the benchmark's standardized format:
\begin{itemize}
    \item \textit{Introduction to Probability} \citep{grinstead1997introduction}
    \item \textit{Learning Statistics with jamovi: A Tutorial for Beginners in Statistical Analysis} \citep{navarro2022learning}
    \item \textit{Linear Regression Using R: An Introduction to Data Modeling} \citep{lilja2016linear}
    \item \textit{Think Bayes: Bayesian Statistics in Python} \citep{downey2021think}
\end{itemize}

\paragraph{Existing Benchmarks}
We incorporated questions from the following open-source benchmarks:
\begin{itemize}
    \item \textbf{StatQA} \citep{zhu2024statqa}: 91 questions (License: GPL-3.0).
    \item \textbf{CLADDER} \citep{jin2024cladderassessingcausalreasoning}: 11 causal reasoning questions (License: MIT).
    \item \textbf{CORR2CAUSE} \citep{jin2024largelanguagemodelsinfer}: 11 correlation-to-causation questions (License: MIT).
\end{itemize}

\paragraph{Online Resources}
We supplemented the dataset with real-world materials from public repositories and educational platforms:
\begin{itemize}
    \item \textbf{Kaggle}\footnote{\url{https://www.kaggle.com/}}: Utilized real-world datasets to construct data-intensive tool-use questions (Licenses: CC0 / Open Access).
    \item \textbf{LibreTexts}\footnote{\url{https://stats.libretexts.org/}}: An open-access educational library. We adapted \textbf{statistical exercises and conceptual problems} for decision-making tasks (License: CC-BY-NC-SA).
\end{itemize}

\subsection{Stat-Open}
Stat-Open focuses on evaluating complex problem-solving skills through authentic, open-ended tasks. We curate 30 high-quality problems from established mathematical and statistical modeling competitions.

\paragraph{Sources.}
The Stat-Open tasks are drawn from the following competitions:

\begin{itemize}
    \item \textbf{CUMCM} (China Undergraduate Mathematical Contest in Modeling)\footnote{\url{http://www.mcm.edu.cn/}}: Also known as the ``Higher Education Press Cup''. We selected problems involving comprehensive data analysis and modeling.
    
    \item \textbf{MCM/ICM} (Mathematical Contest in Modeling / International Mathematical Contest in Modeling)\footnote{\url{https://www.comap.com/contests/mcm-icm}}: Organized by COMAP. We focused on problems that require deep statistical reasoning (e.g., Problem C, which typically involves data insights).
    
    \item \textbf{MAS} (National Case Competition for Applied Statistics Postgraduates)\footnote{\url{http://mas.ruc.edu.cn/index.htm}}: Organized by the National Steering Committee for Postgraduate Education of Applied Statistics. These problems are selected for their emphasis on real-world statistical data analysis.
\end{itemize}

\paragraph{Copyright and Release.}
The original problem statements and associated datasets remain the intellectual property of their respective competition organizers. We use these materials for non-commercial academic research and preserve their original ownership while standardizing file formats to facilitate automated LLM evaluation. Our GPL-3.0 release applies only to the benchmark code, evaluation pipeline, metadata, and other components created by us, and does not relicense the original third-party problem statements or datasets.

\section{Experiment Details}
\label{app:exp_easy}
We utilized a hybrid deployment strategy. Meta-Llama-3.1-8B-Instruct was deployed locally on 8$\times$NVIDIA RTX 4090 GPUs with vLLM \citep{kwon2023efficientmemorymanagementlarge} and the max-model-len parameter was set to 128,000 tokens. All other models (e.g., Qwen2.5 series, DeepSeek-V3, GPT-4o-mini) were accessed via APIs. We set the temperature to 0 for all evaluations. For \textbf{Qwen3-8B}, we explicitly disabled the \texttt{enable\_thinking} option to maintain output format consistency with other instruction-tuned models. For each task, all of frameworks are permitted a maximum of 5 tool calls per interaction turn(In practice,no more than two are ever required). 

\section{Error Analysis}
\label{sec:appendix_errors}
Analysis of 38 practical application tasks where DeepSeek-V3 failed but Qwen2.5-72B succeeded:

\begin{table}[h]
\centering
\resizebox{\columnwidth}{!}{
    \begin{tabular}{l r l}
    \toprule
    \textbf{Category} & \textbf{Count} & \textbf{Issues} \\
    \midrule
    Tool Selection & 19 (50\%) & Incorrect/missing tool invocation \\
    Compliance & 10 (26\%) & Format(6), parameter(3), manual calc(1) \\
    Result Interp. & 9 (24\%) & Correct execution, wrong conclusion \\
    \bottomrule
    \end{tabular}}
\caption{Failure analysis of DeepSeek-V3 on practical application tasks.}
\label{tab:error_analysis}
\end{table}

\section{Licensing}
Similar to StatQA, we also release StatABench under GPL-3. The details for the datasets used in this work are listed below:

\begin{itemize}
    \item \textbf{StatQA}: Licensed under  \href{https://www.gnu.org/licenses/gpl-3.0.en.html}{GPL-3.0}.
    \item \textbf{CLADDER}: Licensed under the \href{https://opensource.org/licenses/MIT}{MIT License}.
    \item \textbf{CORR2CAUSE}: Licensed under the \href{https://opensource.org/licenses/MIT}{MIT License}.
\end{itemize}

\section{Prompts}
We list all the prompts that we used in this work.

\begin{figure*}[h]
    \includegraphics[width=\textwidth]{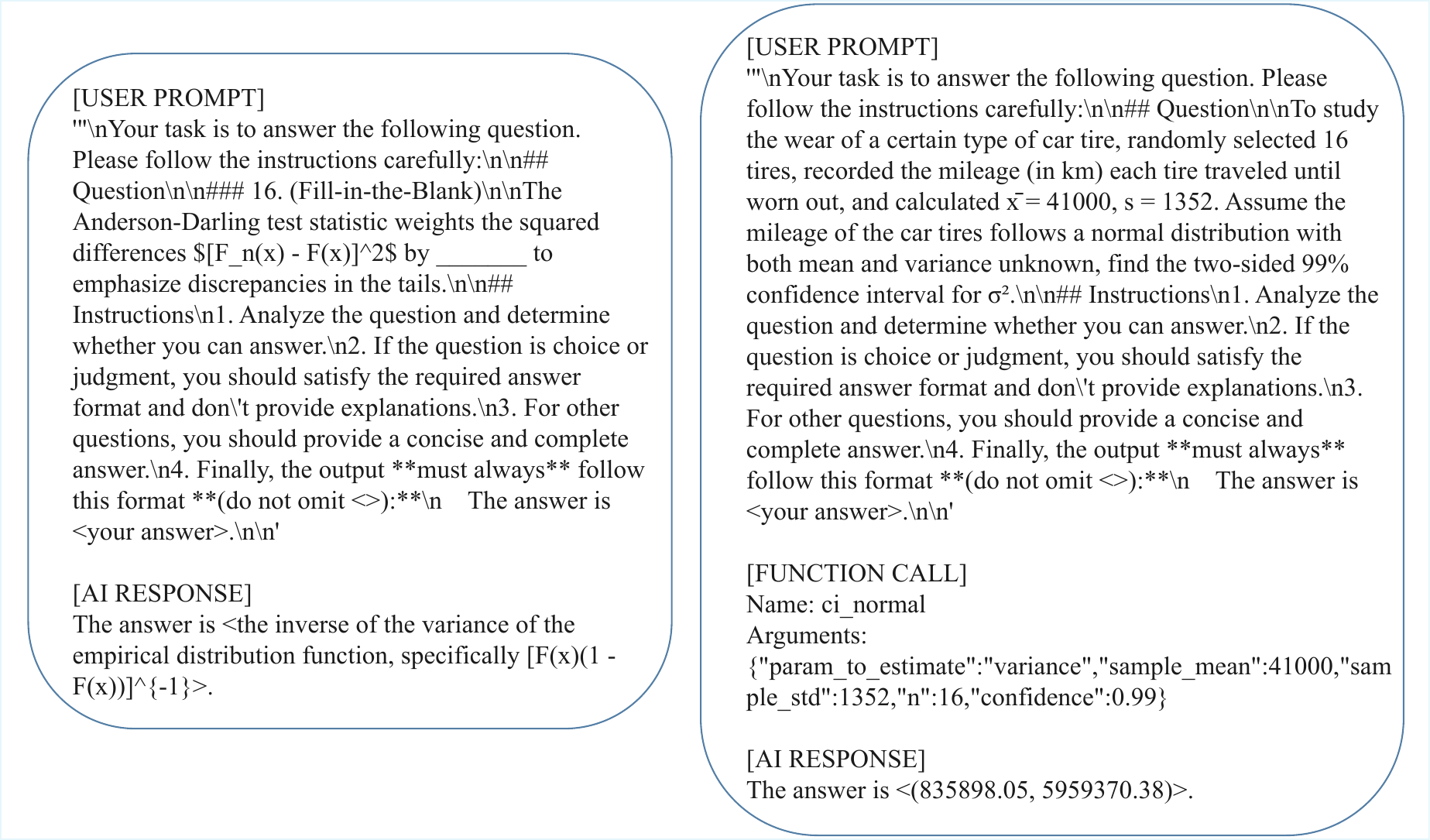}
    \caption{Comparison between fundamental statistical task (left) and practical application task (right) workflows. Both questions are answerd by DeepSeekV3 through LangChain MCP framework.}
    \label{fig:ex_ques1}
\end{figure*}

\begin{figure*}[t]
    \centering
    \includegraphics[width=\textwidth]{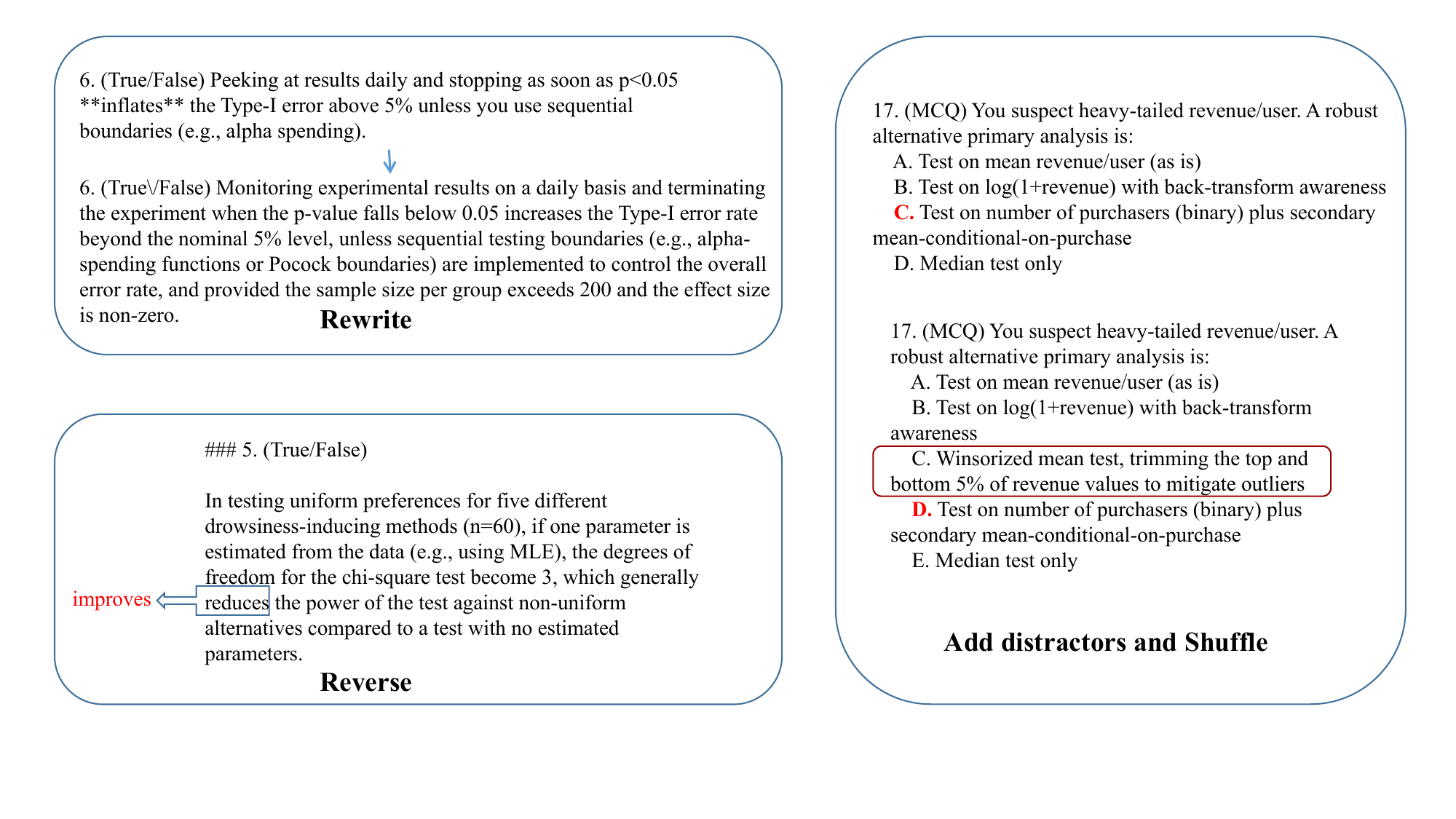}
    \caption{\textbf{Visualization of the Modify methods.} As detailed in the Decontamination section, we apply rigorous perturbations to flagged instances to mitigate potential data leakage. The strategies include: (1) \textbf{Rewriting} (top-left), where problem stems are paraphrased to alter surface patterns; (2) \textbf{Logical Inversion} (bottom-left), corresponding to the ``inverting True/False conditions'' strategy where key predicates are reversed; and (3) \textbf{Distractor Injection \& Shuffling} (right), which expands the option set to prevent memorization-based shortcuts. These manual interventions ensure the benchmark evaluates robust statistical understanding.}
    \label{fig:decom}
\end{figure*}

\begin{table*}[h]
\centering
\resizebox{\textwidth}{!}{
    \begin{tabular}{l r l}
    \toprule
    \textbf{Error Category} & \textbf{Count} & \textbf{Specific Issues} \\
    \midrule
    \textbf{Tool Selection Failure} & \textbf{19 (50\%)} & Invoking incorrect tools or failing to invoke required tools. \\
    \textbf{Compliance \& Execution} & 10 (26\%) & Answer format error (6), Parameter error (3), Manual calculation (1). \\
    \textbf{Result Interpretation} & 9 (24\%) & Correct tool execution but incorrect or unsupported statistical conclusion. \\
    \midrule
    \textbf{Total Mismatch} & 38 & Cases: Qwen2.5-72B (\ding{51}) vs. DeepSeek-V3 (\ding{53}). \\
    \bottomrule
    \end{tabular}}
\caption{\textbf{Failure Analysis of DeepSeek-V3 on practical application Tasks.} We analyze 38 instances where DeepSeek-V3 failed while Qwen2.5-72B succeeded. The dominance of selection and compliance errors (76\%) suggests the rigid baseline framework restricts DeepSeek's potential in tool invocation.}
\label{tab:error_analysis}
\end{table*}

\begin{table*}[t]
\centering
\renewcommand{\arraystretch}{1.1} 
\label{tab:functions_part1}
\resizebox{\textwidth}{!}{
\begin{tabular}{p{5.5cm} p{10cm}}
    \toprule
    \textbf{Function Name} & \textbf{Description} \\
    \midrule
    
    \texttt{calculate\_statistic} & 
    Computes descriptive statistics (e.g., mean, median, skewness, kurtosis) for a specified data column. \\
    
    \texttt{check\_missing\_values} & 
    Analyzes the count and percentage of missing values (NaNs) in specified columns. \\
    
    \texttt{detect\_outliers} & 
    Identifies outliers in numeric columns using the Interquartile Range (IQR) method. \\
    
    \texttt{check\_column\_type\_is} & 
    Validates whether specified columns match a target data type (numeric, object, or datetime). \\
    
    \texttt{correlation\_analysis} & 
    Computes Pearson, Kendall, or partial correlation matrices for numeric variables. \\
    
    \texttt{ci\_normal} & 
    Calculates confidence intervals for the mean or variance of a normal population (one-sample). \\
    
    \texttt{ci\_two\_normal} & 
    Computes confidence intervals for the difference in means or ratio of variances between two samples. \\
    
    \texttt{contingency\_test} & 
    Performs statistical tests on contingency tables, including Chi-square, Fisher's Exact, and Mantel-Haenszel tests. \\
    
    \texttt{ks\_test} & 
    Conducts one-sample or two-sample Kolmogorov-Smirnov tests to check goodness-of-fit. \\
    
    \texttt{simple\_linear\_regression} & 
    Fits a simple Ordinary Least Squares (OLS) regression model and reports coefficients and $R^2$. \\
    
    \texttt{mood\_variance\_test} & 
    Performs Mood's test to compare the scale parameters (variances) of two distributions. \\
    
    \texttt{multivariable\_linear\_regression} & 
    Fits a multivariable OLS model and calculates Variance Inflation Factors (VIF) for collinearity checks. \\
    
    \texttt{run\_glm} & 
    Fits Generalized Linear Models (GLM) supporting various families (Gaussian, Poisson, Logistic, etc.). \\
    
    \texttt{nonparametric\_test} & 
    Executes nonparametric tests including Mann–Whitney U, Wilcoxon Signed-Rank, and Kruskal–Wallis. \\
    
    \texttt{huber\_regression} & 
    Fits a robust linear regression model using Huber loss to reduce sensitivity to outliers. \\
    
    \texttt{test\_stationarity} & 
    Tests time-series stationarity using Augmented Dickey-Fuller (ADF) or KPSS tests. \\
    
    \texttt{decompose\_stl} & 
    Decomposes a time series into trend, seasonal, and residual components using LOESS (STL). \\
    
    \texttt{auto\_arima\_modeling} & 
    Automatically selects and fits the optimal ARIMA model based on information criteria (AIC/BIC). \\
    
    \bottomrule
\end{tabular}}
\caption{Summary of statistical functions in our SAToolkit (Part 1 of 2)}
\end{table*}

\begin{table*}[t]
\centering
\renewcommand{\arraystretch}{1.1}
\resizebox{\textwidth}{!}{
\label{tab:functions_part2}
\begin{tabular}{p{5.5cm} p{10cm}}
    \toprule
    \textbf{Function Name} & \textbf{Description} \\
    \midrule
    
    \texttt{kaplan\_meier\_plot} & 
    Estimates survival functions and timelines using the Kaplan-Meier estimator. \\
    
    \texttt{logrank\_test\_compare} & 
    Performs the Log-rank test to compare survival distributions between different groups. \\
    
    \texttt{fit\_cox\_model} & 
    Fits a Cox Proportional Hazards regression model to analyze survival data with covariates. \\
    
    \texttt{ab\_ttest} & 
    Conducts Welch's t-test for comparing means in A/B testing scenarios. \\
    
    \texttt{bootstrap\_abtest} & 
    Estimates confidence intervals for differences in means or medians using bootstrapping. \\
    
    \texttt{ab\_power\_analysis} & 
    Calculates the required sample size for A/B tests based on desired power and effect size. \\
    
    \texttt{advanced\_regression} & 
    Fits high-dimensional regression models using Lasso, Ridge, or ElasticNet regularization. \\
    
    \texttt{sparse\_pca\_analysis} & 
    Performs Sparse Principal Component Analysis (SPCA) to extract interpretable components. \\
    
    \texttt{bayesian\_inference} & 
    Computes posterior distributions and credible intervals for Beta-Binomial or Normal-Normal models. \\
    
    \texttt{bayesian\_linear\_regression} & 
    Fits a Bayesian Ridge regression model and estimates posterior distributions of coefficients. \\
    
    \texttt{fit\_hierarchical\_model} & 
    Fits hierarchical Bayesian models (Normal or Binomial) using MCMC sampling via PyMC. \\
    
    \texttt{estimate\_ATT\_with\_psm} & 
    Estimates the Average Treatment Effect on the Treated (ATT) using Propensity Score Matching. \\
    
    \texttt{estimate\_did\_effect} & 
    Estimates causal effects using the Difference-in-Differences (DID) method for panel data. \\
    
    \texttt{synthetic\_control} & 
    Constructs a synthetic counterfactual to estimate treatment effects for a single treated unit. \\
    
    \texttt{fdr\_control\_df} & 
    Adjusts p-values to control the False Discovery Rate (FDR) using Benjamini-Hochberg/Yekutieli methods. \\
    
    \texttt{fwer\_control\_df} & 
    Adjusts p-values to control the Family-Wise Error Rate (FWER) using Bonferroni or Holm methods. \\
    
    \texttt{show\_csv\_info\_en} & 
    Displays basic metadata, column names, and a preview of the dataset. \\
    
    \bottomrule
\end{tabular}}
\caption{Summary of statistical functions in SAToolkit (Part 2 of 2)}
\end{table*}

\begin{figure*}[h]
    \centering
    \includegraphics[width=\textwidth,
    height= 0.75\textheight,
    keepaspectratio]{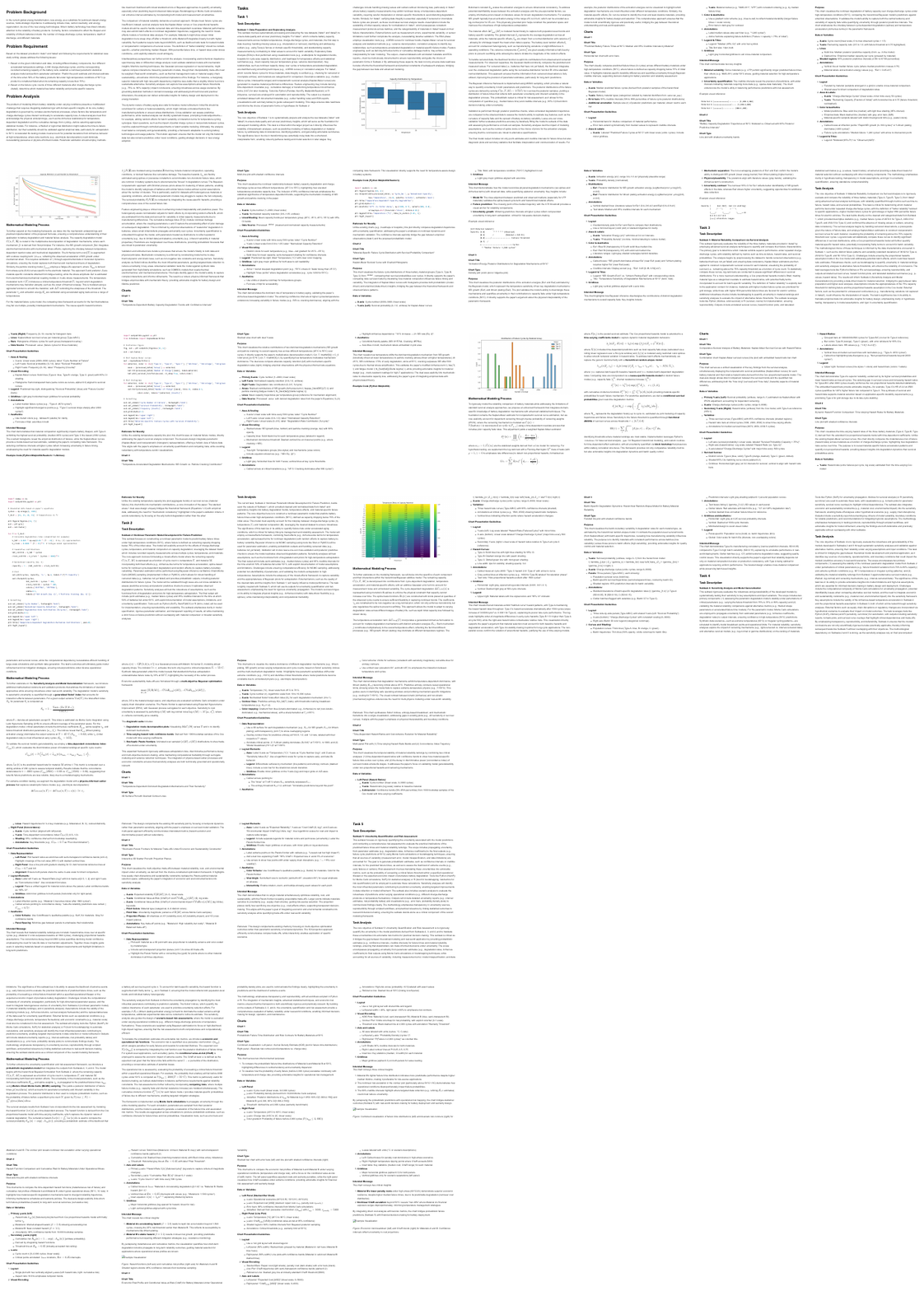}
    \caption{The report generated by LLM-MM-Agent for Problem a of the 2022 MAS}
    \label{fig:hard_report1}
\end{figure*}

\begin{figure*}[h]
    \centering
    \includegraphics[width=\textwidth,
    height= 0.75\textheight,
    keepaspectratio]{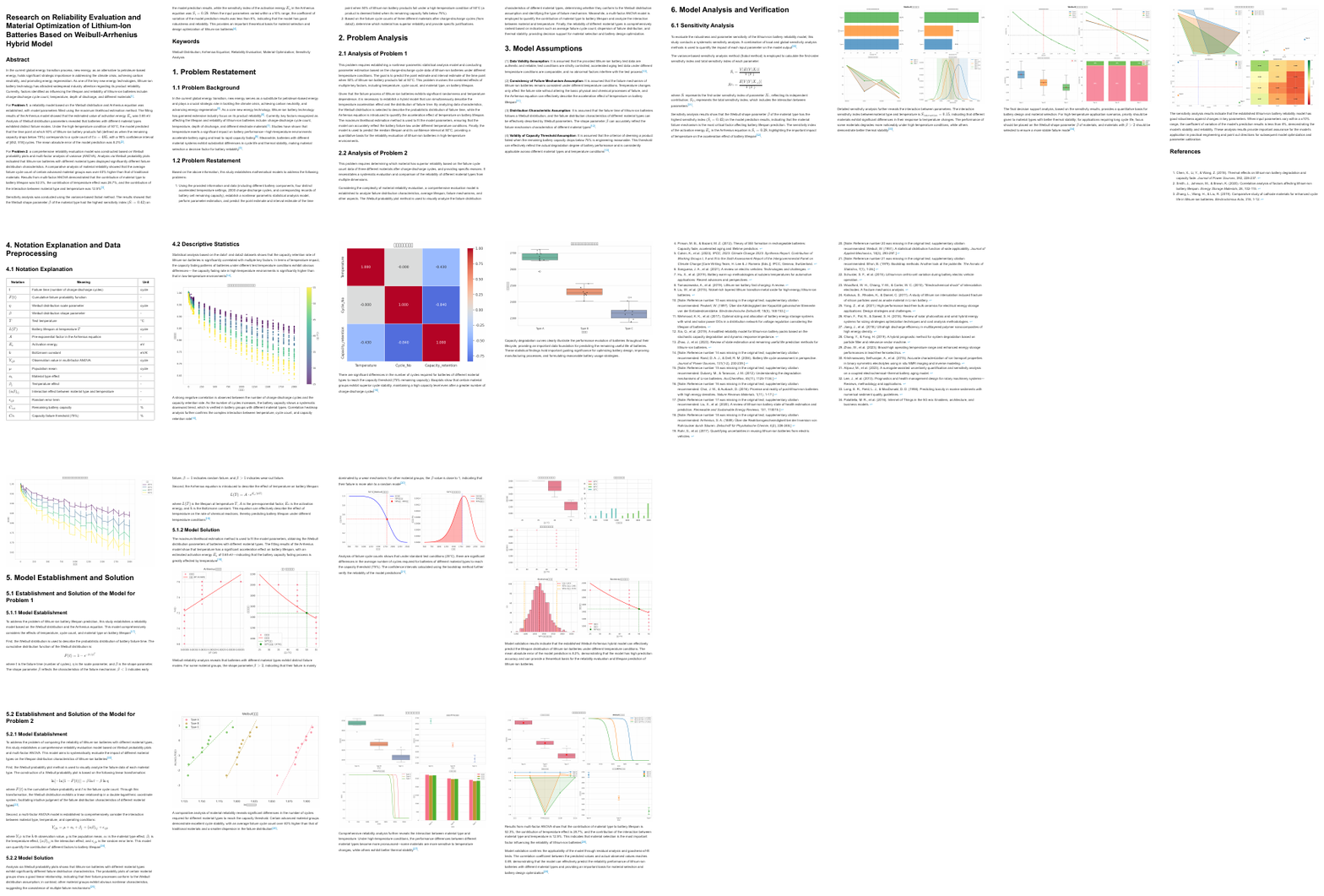}
    \caption{The report generated by MathModelAgent for Problem a of the 2022 MAS}
    \label{fig:hard_report2}
\end{figure*}

\begin{figure*}[htbp]
    \begin{tcolorbox}[
          title=\textbf{Prompt for All questions},
          colback=white,
          colframe=black,
          left=1em,
          right=1em,
          boxsep=0.8mm
        ]
        \footnotesize
        \textbf{[SYSTEM]}\\
        You are a professional statistics analyst, and could answer user's questions with or without using tools.
        
        \vspace{0.6em}
        \textbf{[USER]}\\
        Your task is to answer the following question. Please follow the instructions carefully:
        
        \vspace{0.5em}
        \textbf{Question}\\
        \texttt{\{questions\}}
        
        \vspace{0.5em}
        \textbf{Instructions}
        \begin{enumerate}[leftmargin=1.2em, nosep]
          \item Analyze the question and determine whether you can answer.
          \item If the question is choice or judgment, satisfy the required answer format and do not provide explanations.
          \item For other questions, provide a concise and complete answer.
          \item Finally, the output \textbf{must always} follow this format (do not omit angle brackets):\\
          \texttt{The answer is <your answer>.}
        \end{enumerate}
    \end{tcolorbox}
    \caption{The prompt used in the LangChain MCP Frame}
\end{figure*}

\begin{figure*}[htbp]
    \begin{tcolorbox}[
          title=\textbf{Prompt for Comparing Answers},
          colback=white,
          colframe=black,
          left=1em,
          right=1em,
          boxsep=0.8mm
        ]
        \footnotesize
        For the following question:\\
        \texttt{\{question\}}
        
        \vspace{0.5em}
        There are two answers:\\
        \texttt{<ground truth> \{ground\} </ground truth>}\\
        \texttt{<response> \{response\} </response>}
        
        \vspace{0.6em}
        Please determine whether the response correctly conveys the meaning of the ground. You should follow these rules:
        \begin{itemize}[leftmargin=1.2em, nosep]
          \item A response should be considered correct ONLY IF it either:
          \begin{itemize}[leftmargin=1.2em, nosep]
            \item explicitly states a conclusion that answers the question (e.g., whether a difference exists or not), and this conclusion aligns with the ground truth; OR
            \item provides explicit and sufficient results (e.g., test statistics, p-values, model outputs) from which the correct conclusion can be directly and unambiguously determined, and these results are consistent with the ground truth.
          \end{itemize}
          \item If the response does not explicitly answer the question with a conclusion, and instead only discusses methodology, analysis plans, or conditional statements (e.g., ``if significant, then...''), it must be judged as false.
          \item If the main conclusion is wrong, missing, or includes factual information that contradicts the ground truth, output ``false''.
        \end{itemize}
        
        \vspace{0.5em}
        \textbf{Important:} Your answer must be strictly ``true'' or ``false'' only, without any additional explanation.
    \end{tcolorbox}
    \caption{The prompt for verifying the model's response}
\end{figure*}

\begin{figure*}[htbp]
    \begin{tcolorbox}[
          title=\textbf{Prompt for Rewrite Final Answer},
          colback=white,
          colframe=black,
          left=1em,
          right=1em,
          boxsep=0.8mm
        ]
        \footnotesize

        I have a question--answer pair, where the question is as follows:\\
        \texttt{\{question\}}
        
        \vspace{0.5em}
        The original reference answer consists of two parts:
        \begin{enumerate}[leftmargin=1.2em, nosep]
          \item The function call string\\
          \texttt{\{user\_code\}}
          \item The execution result after calling the function\\
          \texttt{\{code\_result\}}
        \end{enumerate}
        
        \vspace{0.6em}
        Based on this information, generate a standardized answer that directly addresses the question.
        
        \vspace{0.5em}
        \textbf{Requirements:}
        \begin{itemize}[leftmargin=1.2em, nosep]
          \item The answer should be fluent, concise, and rigorous, and should correctly answer the question.
          \item The output must be in English.
          \item The answer should be brief and focus on the final result, without unnecessary explanation of the solution process.
          \item Ambiguity should be minimized to facilitate downstream evaluation by other models.
        \end{itemize}
        
        \vspace{0.5em}
        Please output only the rewritten standard answer, without any additional explanation or commentary.
    \end{tcolorbox}
    \caption{The prompt for generating final new answer }
\end{figure*}

\begin{figure*}[htbp]
    \begin{tcolorbox}[
          title=\textbf{Generate Questions Prompt 1},
          colback=white,
          colframe=black,
          left=1em,
          right=1em,
          boxsep=0.8mm
        ]
        \footnotesize
        You are given a dataset and a function as follows:

        \vspace{0.4em}
        \texttt{\{function\}}

        \vspace{0.6em}
        Your task:
        \begin{enumerate}[leftmargin=1.2em, nosep]
          \item Create several meaningful problems that can be solved directly using this function, based on the dataset.
          \item Each problem must be solvable by calling this function exactly once, without using any additional helper functions.
          \item For each problem, provide:
          \begin{itemize}[leftmargin=1.2em, nosep]
            \item A clear problem description (what needs to be solved).
            \item The exact function call (showing how to use the function to solve the problem).
            \item A short analysis explaining why this function can solve the problem and how it works in this context.
          \end{itemize}
        \end{enumerate}

        \vspace{0.4em}
        The goal is to generate diverse, realistic questions that demonstrate how this function can be applied to the dataset.
    \end{tcolorbox}
    \caption{The prompt for generating practical application questions (Path I)}
\end{figure*}

\begin{figure*}[htbp]
    \begin{tcolorbox}[
          title=\textbf{Generate Questions Prompt 2},
          colback=white,
          colframe=black,
          left=1em,
          right=1em,
          boxsep=0.8mm
        ]
        \footnotesize
        I would like you to help me create problems and corresponding datasets based on the following function. For each problem, there must be a dataset that can be used to solve it by calling this function exactly once, without relying on any additional helper functions.

        \vspace{0.4em}
        \texttt{\{function\}}

        \vspace{0.6em}
        Requirements:
        \begin{enumerate}[leftmargin=1.2em, nosep]
          \item Each problem should be solvable by a single call to the given function.
          \item For each problem, clearly show how to call this function to obtain the correct result.
          \item Provide Python code to generate the dataset used in the problem, ensuring that the dataset matches the problem description.
          \item Different problems may share the same dataset.
          \item Include an analytical explanation of why this function can be used to solve the given problem.
        \end{enumerate}
    \end{tcolorbox}
    \caption{The prompt for generating practical application questions (Path II)}
\end{figure*}

\begin{figure*}[htbp]
    \begin{tcolorbox}[
        title=\textbf{Evaluate Structural Coherency Prompt},
        colback=white,
        colframe=black,
        left=1em,
        right=1em,
        boxsep=0.8mm,
        sharp corners,
        fontupper=\footnotesize
        ]
        \footnotesize
        \textbf{[SYSTEM]}\\
        You are an expert judge evaluating statistical modeling papers. Your task is to assess the structural coherency of the paper by checking if it contains all necessary components.
        
        \vspace{0.4em}
        Key components to evaluate (up to 1 point each):
        
        \vspace{0.4em}
        \textbf{1. Problem Restatement (0-1):}\\
        0.00: Missing or completely misunderstood\\
        ...
        
        \vspace{0.4em}
        \textbf{2. Assumptions and Justification (0-1):}\\
        0.00: Missing or unjustified\\
        \hspace*{1em} Example: No assumptions listed or completely unreasonable ones\\
        ...
        
        \vspace{0.4em}
        \textbf{3. Modeling Implementation (0-1):}\\
        0.00: Missing or fundamentally flawed\\
        \hspace*{1em} Example: No clear mathematical formulation\\
        ...
        
        \vspace{0.4em}
        \textbf{4. Solution Process (0-1):}\\
        0.00: Missing or invalid\\
        \hspace*{1em} Example: No solution method or completely incorrect approach\\
        ...
        
        \vspace{0.4em}
        \textbf{5. Analysis (0-1):}\\
        0.00: Missing or invalid\\
        \hspace*{1em} Example: No analysis or completely wrong interpretations\\
        ...
        
        \vspace{0.6em}
        Your response must follow this exact format:
        
\begin{verbatim}
Your Response:
```json
{
    "scores": {
        "problem_restatement": 0.0,
        "assumptions": 0.0,
        "modeling_implementation": 0.0,
        "solution_process": 0.0,
        "analysis": 0.0
    },
    "explanation": {
        "problem_restatement": "why this score",
        "assumptions": "why this score",
        "modeling_implementation": "why this score",
        "solution_process": "why this score",
        "analysis": "why this score"
    }
}
\end{verbatim}

    \vspace{0.6em}
    Note: For each component, score must be exactly 0.0, 0.25, 0.50, 0.75, or 1.00. Be extremely critical - most solutions should score in the 0.25-0.50 range unless truly exceptional.
    
    \vspace{0.8em}
    \textbf{[USER]}\\
    Please evaluate the structural coherency of the following statistical modeling paper:
    
    \vspace{0.4em}
    \texttt{\{writing\}}
    
    \vspace{0.4em}
    Provide scores and explanations for each component.
    
    \vspace{0.4em}
    Your Response:
\end{tcolorbox}
\caption{The simplified prompt template used for evaluating structural coherency of the report.}
\label{fig:coherency_prompt}
\end{figure*}

\begin{figure*}[htbp]
    \begin{tcolorbox}[
        title=\textbf{Evaluate Modeling Groundness Prompt},
        colback=white,
        colframe=black,
        left=1em,
        right=1em,
        boxsep=0.8mm,
        sharp corners,
        fontupper=\footnotesize
        ]
        \footnotesize
        \textbf{[SYSTEM]}\\
        You are currently evaluating statistical modeling papers. Your task is to assess how well the solution's modeling approach is grounded in statistical theory and scientific principles. You should evaluate based on the role you are given.
        
        \vspace{0.4em}
        Score each aspect from 0--1, starting at 0 and requiring justification for any increase:
        
        \vspace{0.4em}
        \textbf{1. Statistical Theory and Model Specification (0--1):}\\
        0.00: Fundamentally flawed or missing\\
        Example: No clear probabilistic model, used incorrect statistical concepts for the data type.\\
        ...
        
        \vspace{0.4em}
        \textbf{2. Data Quality and Feature Engineering (0--1):}\\
        0.00: No connection to data reality\\
        Example: Model is presented without any description of the data source, its collection process, or its characteristics.\\
        ...
        
        \vspace{0.4em}
        \textbf{3. Methodology and Estimation (0--1):}\\
        0.00: Elementary/inappropriate\\
        Example: Using methods for independent data on time-series data; using linear models for clearly non-linear, non-transformable relationships.\\
        ...
        
        \vspace{0.4em}
        \textbf{4. Model Validation and Diagnostics (0--1):}\\
        0.00: No validation\\
        Example: Model results are presented without any form of verification or performance metric.\\
        ...
        
        \vspace{0.4em}
        \textbf{5. Implementation Quality and Reproducibility (0--1):}\\
        0.00: Poor/incorrect\\
        Example: Code contains clear errors in statistical formulas or misuse of libraries.\\
        ...
        
        \vspace{0.6em}
        Your response must follow this exact format:
        
\begin{verbatim}
Your Response:
{
    "statistical_theory_and_model_specification": {
        "score": 0.0,
        "explanation": "Detailed justification for score"
    },
    "data_quality_and_feature_engineering": {
        "score": 0.0,
        "explanation": "Detailed justification for score"
    },
    "methodology_and_estimation": {
        "score": 0.0,
        "explanation": "Detailed justification for score"
    },
    "model_validation_and_diagnostics": {
        "score": 0.0,
        "explanation": "Detailed justification for score"
    },
    "implementation_quality_and_reproducibility": {
        "score": 0.0,
        "explanation": "Detailed justification for score"
    },
    "calculated_overall": 0.0,
    "overall_feedback": "Critical analysis of strengths and weaknesses"
}
\end{verbatim}
        
        \vspace{0.6em}
        Note: Scores must be exactly 0.00, 0.25, 0.50, 0.75, or 1.00. Start at 0 and justify each increment. Be extremely critical.
        
        \vspace{0.8em}
        \textbf{[USER]}\\
        Please evaluate the modeling groundedness of the following statistical modeling paper:
        
        \vspace{0.4em}
        \texttt{\{writing\}}
        
        \vspace{0.4em}
        Provide scores and detailed justification for each aspect. Remember your role as \texttt{\{role\_name\}}. Your judgement should be based on this role's perspective.
        
        \vspace{0.4em}
        Your Response:
    \end{tcolorbox}
    \caption{The simplified prompt template for evaluating modeling groundness of the report.}
\end{figure*}

\begin{figure*}[htbp]
    \begin{tcolorbox}[
        enhanced,
        title=\textbf{Evaluate Data Groundedness Prompt},
        colback=white,
        colframe=black,
        left=1em,
        right=1em,
        boxsep=0.8mm,
        sharp corners,
        fontupper=\footnotesize
        ]
        \footnotesize
        \textbf{[SYSTEM]}\\
        You are currently evaluating statistical modeling papers. Your task is to assess how well the solution is grounded in data and evidence. You should evaluate based on the role you are given.
        
        \vspace{0.4em}
        Score each aspect from 0-1, starting at 0 and requiring justification for any increase:
        
        \vspace{0.4em}
        \textbf{1. Data Quality (0-1):}\\
        0.00: No data or invalid data\\
        \hspace*{1em} Example: Made-up numbers without sources\\  
        ...
        
        \vspace{0.4em}
        \textbf{2. Data Processing (0-1):}\\
        0.00: No processing/invalid\\
        \hspace*{1em} Example: Raw data used without cleaning\\
        ...
        
        \vspace{0.4em}
        \textbf{3. Statistical Analysis (0-1):}\\
        0.00: No analysis/incorrect\\
        \hspace*{1em} Example: No statistical methods used\\
        ...
        
        \vspace{0.4em}
        \textbf{4. Data Integration (0-1):}\\
        0.00: No integration\\
        \hspace*{1em} Example: Data disconnected from model\\
        ...
        
        \vspace{0.4em}
        \textbf{5. Validation \& Testing (0-1):}\\
        0.00: No validation\\
        \hspace*{1em} Example: Results accepted without testing\\
        ...
        
        \vspace{0.6em}
        Your response must follow this exact format:
        
\begin{verbatim}
Your Response:
```json
{
    "data_quality": {
        "score": 0.0,
        "explanation": "Detailed justification for score"
    },
    "data_processing": {
        "score": 0.0,
        "explanation": "Detailed justification for score"
    },
    ...,
    "validation": {
        "score": 0.0,
        "explanation": "Detailed justification for score"
    },
    "calculated_overall": 0.0,
    "overall_feedback": "Critical analysis of strengths and weaknesses"
}
\end{verbatim}

    \vspace{0.6em}
    Note: Scores must be exactly 0.00, 0.25, 0.50, 0.75, or 1.00. Start at 0 and justify each increment. Be extremely critical. You should also give your score and explaination from your role's perspective.
    
    \vspace{0.8em}
    \textbf{[USER]}\\
    Please evaluate the data groundedness of the following statistical modeling paper:
    
    \vspace{0.4em}
    \texttt{\{writing\}}
    
    \vspace{0.4em}
    Provide scores and detailed justification for each aspect. Remember your role as \texttt{\{role\_name\}}. Your judgement should be based on this role's perspective.
    
    \vspace{0.4em}
    Your Response:
\end{tcolorbox}
\caption{The simplified prompt template used for evaluating data groundedness of the report.}
\label{fig:groundedness_prompt}
\end{figure*}

\begin{figure*}[htbp]
    \begin{tcolorbox}[
        enhanced,
        title=\textbf{Evaluate Analysis Groundedness Prompt},
        colback=white,
        colframe=black,
        left=1em,
        right=1em,
        boxsep=0.8mm,
        sharp corners,
        fontupper=\footnotesize
        ]
        \footnotesize
        \textbf{[SYSTEM]}\\
        You are currently evaluating statistical modeling papers. Your task is to assess the depth and rigor of the analysis derived from the statistical model. You should evaluate based on the role you are given.
        
        \vspace{0.4em}
        Score each aspect from 0-1, starting at 0 and requiring justification for any increase:
        
        \vspace{0.4em}
        \textbf{1. Depth of Statistical Analysis (0-1):}\\
        0.00: No meaningful analysis\\
        ...
        
        \vspace{0.4em}
        \textbf{2. Statistical Rigor and Justification (0-1):}\\
        0.00: No statistical support\\
        \hspace*{1em} Example: Makes claims about trends or differences without any statistical tests or evidence.\\
        ...
        
        \vspace{0.4em}
        \textbf{3. Interpretation of Results and Context (0-1):}\\
        0.00: No interpretation\\
        \hspace*{1em} Example: Presents a table of results or a plot with no explanation.\\
        ...
        
        \vspace{0.4em}
        \textbf{4. Critical Evaluation of Findings (0-1):}\\
        0.00: No critical thinking\\
        \hspace*{1em} Example: Accepts all model outputs as absolute truth without question.\\
        ...
        
        \vspace{0.4em}
        \textbf{5. Implications and Future Directions (0-1):}\\
        0.00: No discussion\\
        \hspace*{1em} Example: The paper ends abruptly after presenting the results.\\
        ...
        
        \vspace{0.6em}
        Your response must follow this exact format:
        
\begin{verbatim}
Your Response:
```json
{
    "depth_of_statistical_analysis": {
        "score": 0.0,
        "explanation": "Detailed justification for score"
    },
    "statistical_rigor_and_justification": {
        "score": 0.0,
        "explanation": "Detailed justification for score"
    },
...,
    "implications_and_future_directions": {
        "score": 0.0,
        "explanation": "Detailed justification for score"
    },
    "calculated_overall": 0.0,
    "overall_feedback": "Critical analysis of strengths and weaknesses"
}

```

\end{verbatim}

```
    \vspace{0.6em}
    Note: Scores must be exactly 0.00, 0.25, 0.50, 0.75, or 1.00. Start at 0 and justify each increment. Be extremely critical. You should also give your score and explanation from your role's perspective.
    
    \vspace{0.8em}
    \textbf{[USER]}\\
    Please evaluate the analysis groundedness of the following statistical modeling paper:
    
    \vspace{0.4em}
    \texttt{\{writing\}}
    
    \vspace{0.4em}
    Provide scores and detailed justification for each aspect. Remember your role as \texttt{\{role\_name\}}. Your judgement should be based on this role's perspective.
    
    \vspace{0.4em}
    Your Response:
\end{tcolorbox}
\caption{The simplified prompt template for evaluating analysis groundedness of the report.}
\label{fig:analysis_prompt}
\end{figure*}

\begin{figure*}[htbp]
    \begin{tcolorbox}[
        enhanced,
        title=\textbf{Evaluate Innovativeness Prompt},
        colback=white,
        colframe=black,
        left=1em,
        right=1em,
        boxsep=0.8mm,
        sharp corners,
        fontupper=\footnotesize
        ]
        \footnotesize
        \textbf{[SYSTEM]}\\
        You are currently evaluating statistical modeling papers. Your task is to assess the innovativeness and originality of the solution approach. You should evaluate based on the role you are given.
        
        \vspace{0.4em}
        Score each aspect from 0--1, starting at 0 and requiring justification for any increase:
        
        \vspace{0.4em}
        \textbf{1. Methodological Innovation (0--1):}\\
        0.00: Standard/textbook approach\\
        \hspace*{1em} Example: Using basic linear regression without modification\\
        ...
        
        \vspace{0.4em}
        \textbf{2. Problem Framing (0--1):}\\
        0.00: Conventional perspective\\
        \hspace*{1em} Example: Following typical problem formulation\\
        ...
        
        \vspace{0.4em}
        \textbf{3. Solution Creativity (0--1):}\\
        0.00: Standard solution\\
        \hspace*{1em} Example: Direct application of known methods\\
        ...
        
        \vspace{0.4em}
        \textbf{4. Technical Advancement (0--1):}\\
        0.00: No advancement\\
        \hspace*{1em} Example: Uses only existing techniques\\
        ...
        
        \vspace{0.4em}
        \textbf{5. Impact Potential (0--1):}\\
        0.00: Minimal impact\\
        \hspace*{1em} Example: No new insights or applications\\
        ...
        
        \vspace{0.6em}
        Your response must follow this exact format:
        
\begin{verbatim}
Your Response:
```json
{
    "methodological_innovation": {
        "score": 0.0,
        "explanation": "Detailed justification for score"
    },
    "problem_framing": {
        "score": 0.0,
        "explanation": "Detailed justification for score"
    },
    ...,
    "impact_potential": {
        "score": 0.0,
        "explanation": "Detailed justification for score"
    },
    "overall_score": 0.0,
    "overall_feedback": "Critical analysis of innovative aspects and potential impact"
}

```

\end{verbatim}

    \vspace{0.6em}
    Note: Scores must be exactly 0.00, 0.25, 0.50, 0.75, or 1.00. Start at 0 and justify each increment. Be extremely critical - true innovation is rare. You should also give your score and explaination from your role's perspective.
    
    \vspace{0.8em}
    \textbf{[USER]}\\
    Please evaluate the innovativeness of the following statistical modeling paper:
    
    \vspace{0.4em}
    \texttt{\{writing\}}
    
    \vspace{0.4em}
    Provide scores and detailed justification for each aspect. Remember your role as \texttt{\{role\_name\}}. Your judgement should be based on this role's perspective.
    
    \vspace{0.4em}
    Your Response:
\end{tcolorbox}
\caption{The simplified prompt template for evaluating innovativeness of the report.}
\label{fig:innovation_prompt}
\end{figure*}

\begin{figure*}[htbp]
    \begin{tcolorbox}[
        title=\textbf{Evaluat Practicality and Scientific Validity Prompt},
        colback=white,
        colframe=black,
        left=1em,
        right=1em,
        boxsep=0.8mm,
        sharp corners,
        fontupper=\footnotesize
        ]
        \footnotesize
        \textbf{[SYSTEM]}\\
        Your task is to evaluate the rigor and rationality of the given modeling paper in mathematical modeling, particularly focusing on the assumptions and rationality.
        
        \vspace{0.4em}
        \textbf{Evaluation Criteria}:
        
        \vspace{0.4em}
        \textbf{\#\#\# 2. Rigor and Rationality of Modeling}
        
        \vspace{0.4em}
        \textbf{\#\#\#\# 2.1 Assumptions}\\
        Clear and explicit. These assumptions are the foundation of the model and need to be rigorously justified.
        \begin{itemize}[leftmargin=1.2em, topsep=0pt, itemsep=0pt]
            \item Are the model assumptions clearly explained?
            \item Are the assumptions reasonable and consistent with the background of the actual problem?
            \item Is the rationality and impact of the assumptions considered?
        \end{itemize}
        \textbf{Scoring Criteria}:\\
        1--2 = Completely unreasonable; 3--4 = Partially reasonable; 5--6 = Average; 7--8 = Reasonable; 9--10 = Very reasonable.
        
        \vspace{0.4em}
        \textbf{\#\#\#\# 2.2 Rationality}\\
        The rationality of the model is key to evaluation. Evaluation criteria can include: whether an appropriate model is chosen, whether the model can realistically reflect the problem, etc.
        \begin{itemize}[leftmargin=1.2em, topsep=0pt, itemsep=0pt]
            \item Has the model chosen appropriate methods and metrics?
            \item Does the structure of the model scientifically reflect the actual problem?
        \end{itemize}
        \textbf{Scoring Criteria}:\\
        1--2 = Completely unreasonable; 3--4 = Partially reasonable; 5--6 = Average; 7--8 = Reasonable; 9--10 = Very reasonable.
        
        \vspace{0.4em}
        \textbf{Output Format}:\\
        Example:
\begin{verbatim}
\#\#\# 2.1 Assumptions

**Evaluation:**

The assumptions are crucial for model building, but the modeling analysis does
not describe the assumptions in sufficient detail. The rationality and impact of
the assumptions are not fully justified... [omitted for brevity] ...making the
foundation of the model less robust.
**Score:**
<reason> The model assumptions are not clear enough... </reason>  
<score> 3 </score>  
\#\#\# 2.2 Rationality 

**Evaluation:**

The rationality of the model is average... [omitted for brevity]
**Score:**
<reason> The rationality of the model is average... </reason>  
<score> 5 </score>
\end{verbatim}

        \vspace{0.4em}
        Please objectively and detailedly evaluate the rigor and rationality of the modeling according to the above evaluation criteria, and give the final score and reason.
        
        \vspace{0.8em}
        \textbf{[USER]}\\
        Please evaluate the practicality and scientificity of the following statistical modeling paper:
        
        \vspace{0.4em}
        \texttt{\{writing\}}
        
        \vspace{0.4em}
        Provide scores and explanations for each component.
        
        \vspace{0.4em}
        Your Response:
    \end{tcolorbox}
    \caption{The prompt template for evaluating the practicality and scientific validity of the report.}
    \label{fig:rigor_prompt}
\end{figure*}

\begin{figure*}[htbp]
    \begin{tcolorbox}[
        enhanced,
        title=\textbf{Evaluate Result Interpretation and Bias Analysis Prompt},
        colback=white,
        colframe=black,
        left=1em,
        right=1em,
        boxsep=0.8mm,
        sharp corners,
        fontupper=\footnotesize
        ]
        \footnotesize
        \textbf{[SYSTEM]}\\
        Your task is to evaluate the result analysis and bias analysis of the given modeling paper, particularly focusing on the rationality, interpretability of the model output, and the identification and correction of biases.
        
        \vspace{0.4em}
        \textbf{Evaluation Criteria}:
        
        \vspace{0.4em}
        \textbf{\#\#\# 4. Result Analysis and Bias Analysis}
        
        \vspace{0.4em}
        \textbf{\#\#\#\# 4.1 Result Analysis}
        \begin{itemize}[leftmargin=1.2em, topsep=0pt, itemsep=0pt]
            \item Are the model output results clear and as expected?
            \item Does the result provide sufficient analysis to explain the model's inference process?
            \item Are the model results interpretable and do they help in understanding the essence of the problem?
            \item Does the analysis provide clear conclusions and highlight the strengths and weaknesses of the model?
        \end{itemize}
        \textbf{Scoring Criteria}:\\
        1--2 = Completely unclear; 3--4 = Partially clear; 5--6 = Average; 7--8 = Clear; 9--10 = Very clear.
        
        \vspace{0.4em}
        \textbf{\#\#\#\# 4.2 Bias Analysis}
        \begin{itemize}[leftmargin=1.2em, topsep=0pt, itemsep=0pt]
            \item Does the model identify and analyze potential biases?
            \item Does it consider data bias, model bias, and other factors?
            \item Does the model appropriately correct biases to reduce their impact on the results?
        \end{itemize}
        \textbf{Scoring Criteria}:\\
        1--2 = Completely ignored biases; 3--4 = Partially considered biases; 5--6 = Average; 7--8 = Considered biases and corrected; 9--10 = Very thorough, biases effectively corrected.
        
        \vspace{0.4em}
        \textbf{Output Format}:\\
        Example 1:
\begin{verbatim}
\#\#\# 4.1 Result Analysis

**Evaluation:**

The model output results are clear and well explain the model's inference
process. The modeler has detailed the background and significance of the model
results, helping to understand the core of the problem. The results show a
reasonable inference path, making the entire analysis process more transparent.
The analysis also provides clear conclusions and highlights the strengths and
weaknesses of the model.
**Score:**
<reason> The result analysis is very clear and effectively supports
decision-making </reason> 
<score> 9 </score>  

\#\#\# 4.2 Bias Analysis

**Evaluation:**

The model effectively identifies and analyzes biases, particularly potential
data biases. The modeler provides correction measures for biases and explains
how these corrections affect the model results. Although there are still some
biases in certain aspects of the model, overall, a comprehensive correction
has been made.
**Score:**
<reason> The bias analysis is thorough, and biases have been effectively
corrected </reason> 
<score> 8 </score>
\end{verbatim}

        \vspace{0.4em}
        Please objectively and detailedly evaluate the result analysis and bias analysis of the modeling according to the above evaluation criteria, and provide the final score and reason.
        \vspace{0.4em}
        \#\#\# 4.1 Result Analysis\textbackslash n\textbackslash n**Evaluation:
        
        \vspace{0.8em}
        \textbf{[USER]}\\
        Please evaluate the practicality and scientificity of the following statistical modeling paper:
        
        \vspace{0.4em}
        \texttt{\{writing\}}
        
        \vspace{0.4em}
        Provide scores and explanations for each component.
        
        \vspace{0.4em}
        Your Response:
    \end{tcolorbox}
    \caption{The prompt template for evaluating result interpretation and bias analysis of the report.}
    \label{fig:bias_prompt}
    
\end{figure*}

\end{document}